\documentclass{midl} 

\usepackage{mwe} 

\usepackage{siunitx}
\usepackage[T1]{fontenc}
\usepackage{amsmath}
\usepackage{float}
\usepackage{makecell}
\usepackage{multirow}
\usepackage{booktabs} 
\usepackage{color}
\usepackage[normalem]{ulem}
\usepackage{xcolor}
\colorlet{changes}{green!45!black} 

\jmlrvolume{-- nnn}
\jmlryear{2026}
\jmlrworkshop{Full Paper -- MIDL 2026}
\editors{Accepted for publication at MIDL 2026}

\title[nnLandmark]{nnLandmark: A Self-Configuring Method for 3D Medical Landmark Detection}

\midlauthor{\Name{Alexandra Ertl\nametag{$^{1,2,3}$}} \orcid{0009-0000-4572-9565} \Email{alexandra.ertl@dkfz-heidelberg.de}\\
\Name{Stefan Denner\nametag{$^{1,4}$}} \orcid{0009-0008-2302-743X} \Email{stefan.denner@dkfz-heidelberg.de}\\
\Name{Robin Peretzke\nametag{$^{1,2,3}$}} \orcid{0000-0002-6187-3636} \Email{robin.peretzke@dkfz-heidelberg.de}\\
\Name{Shuhan Xiao\nametag{$^{1,4}$}} \orcid{0000-0001-5397-814X} \Email{shuhan.xiao@dkfz-heidelberg.de}\\
\Name{David Zimmerer\nametag{$^{1}$}} \orcid{0000-0002-8865-2171} \Email{d.zimmerer@dkfz-heidelberg.de}\\
\Name{Maximilian Fischer\nametag{$^{1,2}$}} \orcid{0009-0001-8305-6370} \Email{maximilian.fischer@dkfz-heidelberg.de}\\
\Name{Markus Bujotzek\nametag{$^{1,2}$}} \orcid{0009-0006-4550-428X} \Email{markus.bujotzek@dkfz-heidelberg.de}\\
\Name{Xin Yang\nametag{$^{6}$}} \orcid{0000-0003-4653-6524} \Email{}\\
\Name{Peter Neher\nametag{$^{1,3,5}$}} \orcid{0000-0002-5285-7554} \Email{p.neher@dkfz-heidelberg.de}\\
\Name{Fabian Isensee\midljointauthortext{Supervised equally}\nametag{$^{1,7}$}} \orcid{0000-0002-3519-5886} \Email{f.isensee@dkfz-heidelberg.de}\\
\Name{Klaus H. Maier-Hein\midlotherjointauthor\nametag{$^{1,2,3,4,5,7,8,9}$}} \orcid{0000-0002-6626-2463} \Email{k.maier-hein@dkfz-heidelberg.de}\\
\addr $^{1}$ German Cancer Research Center (DKFZ) Heidelberg, Division of Medical Image Computing, Heidelberg, Germany \\
\addr $^{2}$ Medical Faculty Heidelberg, Heidelberg University, Heidelberg, Germany \\
\addr $^{3}$ Pattern Analysis and Learning Group, Department of Radiation Oncology, Heidelberg University Hospital, Heidelberg, Germany \\
\addr $^{4}$ Faculty of Mathematics and Computer Science, Heidelberg University, Heidelberg, Germany \\
\addr $^{5}$ German Cancer Consortium (DKTK), DKFZ, core center Heidelberg, Germany \\
\addr $^{6}$ School of Biomedical Engineering, Shenzhen University Medical School, Shenzhen University, Shenzhen, Guangdong, China\\
\addr $^{7}$ Helmholtz Imaging, DKFZ, Heidelberg, Germany \\
\addr $^{8}$ National Center for Tumor Diseases (NCT), NCT Heidelberg, A Partnership Between DKFZ and The University Medical Center Heidelberg, Heidelberg, Germany \\
\addr $^{9}$ HIDSS4Health, Heidelberg, Germany \\
}

\begin{document}

\maketitle

\begin{abstract}
Landmark detection is central to many medical applications, such as identifying critical structures for treatment planning or defining control points for biometric measurements. However, manual annotation is labor-intensive and requires expert anatomical knowledge. 
While deep learning shows promise in automating this task, fair evaluation and interpretation of methods in a broader context, are hindered by limited public benchmarking, inconsistent baseline implementations, and non-standardized experimentation.
To overcome these pitfalls, we present nnLandmark, a self-configuring framework for 3D landmark detection that combines tailored heatmap generation, loss design, inference logic, and a robust set of hyperparameters for heatmap regression, while reusing components from nnU-Net’s underlying self-configuration and training engine. 
nnLandmark achieves state-of-the-art performance across five public and one private dataset, benchmarked against three recently published methods. 
Its out-of-the-box usability enables training strong landmark detection models on new datasets without expert knowledge or dataset-specific hyperparameter tuning.
Beyond accuracy, nnLandmark provides both a strong, common baseline and a flexible, standardized environment for developing and evaluating new methodological contributions. It further streamlines evaluation across multiple datasets by offering data conversion utilities for current public benchmarks. Together, these properties position nnLandmark as a central tool for advancing 3D medical landmark detection through systematic, transparent benchmarking, enabling to genuinely measure methodological progress.
The code is available on GitHub: \url{https://github.com/MIC-DKFZ/nnLandmark}.
\end{abstract}

\begin{keywords}
3D Medical Landmark Detection, Self-Configuration, Benchmarking.
\end{keywords}

\section{Introduction}
\begin{figure}[t]
    \centering
    \includegraphics[width=430pt]{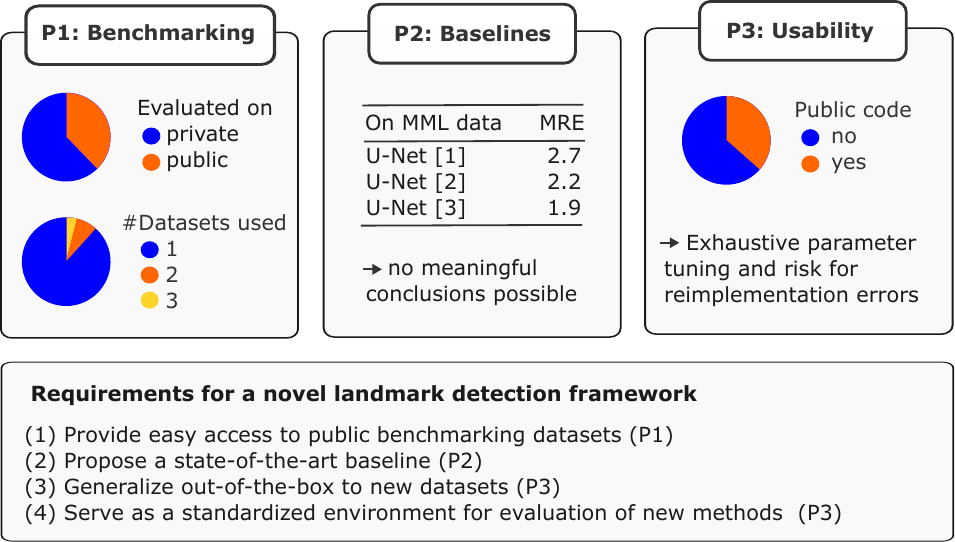}
    \caption{We identified three key pitfalls in the current landmark detection literature regarding benchmarking, baseline comparison and usability, and formulated four practical requirements for a new framework tackling these shortcomings. [1]~\cite{zhang2024SRLD} [2]~\cite{he2024anchor} [3]~\cite{huang2025H3DE}}.  
    \label{fig1}
\end{figure}

The task of medical landmark detection concerns the prediction of coordinates of predefined, anatomical keypoints. 
Accurate localization is critical for several medical imaging applications, including diagnosis, treatment planning, and  navigation. 
In practice, these include the detection of anatomical reference points for image registration, control points for biometric measurements, small critical structures for surgical planning or fetal pose estimation in ultrasound \cite{taha2023magnetic, he2024anchor, Chen2020fetuspose, Gong2025DMGLD}. 
Annotating such keypoints is highly dependent on detailed anatomical knowledge. For example, for fetal brain biometry, this requires reliably localizing the cerebellar landmarks, which is complicated by densely folded cortical structures and low tissue contrast~\cite{Gong2025DMGLD}. 
Further, medical landmark annotation often involves between 10 and 50 landmarks, resulting in a time-consuming process, especially in 3D imaging data. 
Efforts in automating this task based on deep learning have already shown promising results~\cite{schwendicke2021deep,serafin2023accuracy,singh20203d}.
While earlier approaches aimed at directly predicting coordinate values, the current state-of-the-art formulation is heatmap regression. Thereby, each landmark is represented by a Gaussian-like blob in a dedicated output channel. During prediction, the coordinates are derived via channel-wise maximum~\cite{payer2016regressing,pfister2015flowing}.
The default base architecture for pixel-wise heatmap regression is the U-Net \cite{ronneberger2015unet,cciccek20163dunet}. Many efforts have been made to optimize the architecture, aiming to effectively integrate global context or maintain high spatial resolution or super-resolution for sub-pixel accuracy of localization \cite{huang2025H3DE, zhang2024SRLD}. 
However, despite active methodological research in 3D medical landmark detection, progress is still affected by various pitfalls concerning benchmarking and usability of methods (Figure \ref{fig1}), identified from an extensive list of relevant, recent publications (Annex~\ref{sec:app0_fig1}).  
\\
\textbf{Pitfall 1: Insufficient public benchmarking.}
The 3D landmark detection domain suffers from a lack of established and commonly used public benchmarks (Figure \ref{fig1}, left). 
Frequently, new developments only target single, often private datasets, leaving the generalizability of these methods to other tasks in question. This hinders a transparent interpretation of the results in a broader context and limits fair comparison to other methods~\cite{he2024anchor,schwendicke2021deep}. 
While broad public benchmarking is already the standard in segmentation and has greatly propelled the field forward \cite{isensee2021nnu, isensee2024nnu}, in the landmark detection domain, universal insights, generalizable solutions and the broader impact of new developments are often left unexplored by focusing on single and private datasets.
\\
\textbf{Pitfall 2: Inconsistent baseline implementations}
Many publications compare their methods to a 3D U-Net baseline~\cite{ronneberger2015unet,cciccek20163dunet}. However, variations in hyperparameters, implementations, and training setups can substantially change performance, even when using the same underlying architecture. As depicted in Figure \ref{fig1}, center column, this is also evident for landmark detection based on reported U-Net results on the Mandibular Molar Landmark (MML) dataset, with mean radial errors (MRE) ranging from \SIrange{1.9}{2.7}{mm} \cite{huang2025H3DE, he2024anchor, zhang2024SRLD}. 
In absence of a strong, commonly adopted baseline, the field lacks essential context for interpreting the results of new methods and assessing progress across datasets.
%
\\
\textbf{Pitfall 3: Limited out-of-the-box usability.}
The lack of comprehensive benchmarking and the use of non-standardized, custom code bases have led to dataset-specific implementations. Many methods are still published without code or clear instructions on how to adapt them to new datasets, for example when dealing with different modalities or image geometries (Figure \ref{fig1}, right). Applying such methods to new datasets can therefore require substantial expert knowledge in model development and resource-intensive hyperparameter tuning, which complicates broader application and increases the risk of reimplementation errors. The reliance on custom code further introduces potential confounding factors, obscuring the true performance of baseline architectures and leading to unclear conclusions about new developments. A standardized environment that works out-of-the-box across datasets is therefore crucial to enable transparent, systematic evaluation of new methods and quantify true methodological progress.
\\
To counteract these pitfalls, we make the following contributions:
\begin{itemize}
    \item We present a comprehensive benchmark study for 3D medical landmark localization, evaluating three recent state-of-the-art methods across five public and one private datasets that span different imaging modalities and anatomical regions.
    \item We introduce \textbf{nnLandmark}, a fully self-configuring framework for 3D heatmap-based landmark detection that builds on the nnU-Net infrastructure to automatically derive dataset-specific preprocessing and training hyperparameters, enabling robust out-of-the-box generalization to new datasets without manual intervention.
    \item We show that nnLandmark consistently achieves state-of-the-art performance across all six benchmark datasets, surpassing existing methods and establishing a strong, reproducible baseline for future developments in 3D landmark detection.
    \item We demonstrate that nnLandmark serves as a flexible, standardized environment for method development by integrating the H3DE architecture into the framework, yielding clear performance gains over the official implementation and highlighting the value of leveraging a proven experimental infrastructure for systematic ablations and fair evaluation of new methodological contributions.
\end{itemize}

\section{Method}

\begin{figure}[t]
    \centering
    \includegraphics[clip, trim=40pt 170pt 5pt 175pt, width=430pt]{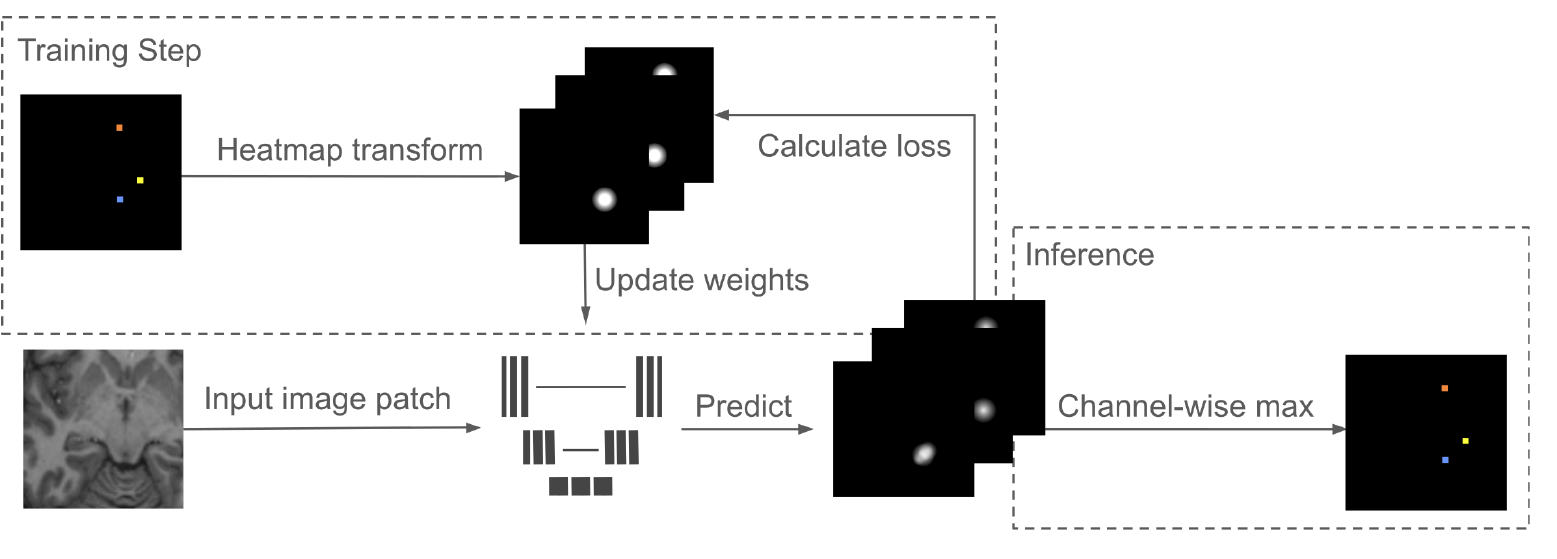}
    \caption{Leveraging nnU-Net's data loading and augmentation pipeline,  landmark segmentation maps are transformed to heatmaps only during loss computation, each landmark represented by a EDT in a dedicated channel. In inference, landmark coordinates are identified by the channel-wise maximum.} 
    \label{fig2}
\end{figure}

Tackling the current pitfalls in landmark detection we derive four practical requirements for a newly proposed framework: (1) Provide easy access to public benchmarking datasets; (2) propose a strong, common baseline, achieving state-of-the-art accuracy across datasets; (3) provide out-of-the-box generalizability for training on new datasets without the need for manual intervention; (4) serve as a flexible and standardized environment for evaluation of new methodological developments.
In segmentation, these requirements have long been understood and are addressed by the well-established nnU-Net framework, which consistently delivers state-of-the-art performance across various datasets~\cite{isensee2021nnu,isensee2024nnu}.
The key concept of nnU-Net is its self-configuration to the task at hand by automatically deriving dataset-specific properties and adjusting preprocessing and hyperparameters for a (residual) U-Net architecture.
Further, nnU-Net has implemented many best practices of image processing for example regarding its data augmentation pipeline and sliding window prediction, which allow translation to heatmap regression.
It is therefore well‑motivated to reuse nnU‑Net as a self‑configuration and training engine. In the following we explore how nnLandmark can be built on this existing infrastructure to arrive at a state-of-the-art, generalizable framework for landmark detection. 
\\
To leverage nnU-Net's data loading pipeline, which expects segmentation inputs, we initially store the landmarks in a multi-label segmentation map, with each landmark represented by a $3 \times 3 \times 3$ voxel label. 
This way nnLandmark can exploit the fully automatic preprocessing and self‑configuration machinery, which has been extensively tuned for 3D medical segmentation. 
The conversion from this multi‑label representation to heatmap regression happens after data augmentation, directly inside the loss computation. For each foreground label, i.e. landmark, the target coordinate is obtained as the center of mass of the corresponding label region. Around this point, an Euclidean distance transform (EDT) with a radius of 15 voxels is injected into a dedicated output channel as the regression target, yielding a smooth distance‑based heatmap (Figure \ref{fig2}). This on‑the‑fly transformation further avoids memory- and CPU-intensive storing and loading of large heatmaps. 
During prediction, a sigmoid activation in the final layer constrains voxel intensities to [0,1], stabilizing  training. Heatmap regression is trained with a Binary Cross‑Entropy (BCE) TopK20 loss, which focuses the gradient signal on the most challenging voxels. Concretely, for every patch the voxel‑wise BCE values are ranked and only the voxels with the highest 20\% loss values contribute to the final loss, which can mitigate the foreground-background imbalance inherent to sparse landmark heatmaps. 
Ablation studies on the size of the EDT and choice of the loss function to evaluate the robustness of our selected parameters can be found in Annex~\ref{sec:app0_ablations}.
For inference, we adapt nnU-Net's sliding window prediction and derive landmark coordinates by taking the channel-wise maximum of the heatmap.

\section{Experiments}

\subsection{Metrics}

The \textbf{Mean Radial Error (MRE)} measures the average Euclidean distance between the predicted and ground truth landmark coordinates. It is defined as:
\begin{equation}
    \text{MRE} = \frac{1}{N} \sum_{i=1}^N ||\mathbf{x}-\mathbf{\hat{x}}||_2
\end{equation}
where $\mathbf{x}$ and $\mathbf{\hat{x}}$ represent the ground truth and predicted coordinates for the $i$-th landmark, respectively, and $N$ is the total number of landmarks. Lower MRE values indicate higher localization accuracy.
\\
The \textbf{Success Detection Rate (SDR)} within a tolerance range quantifies the proportion of detected landmarks that fall within a specified distance threshold $t$ from their ground truth positions, defined as:
\begin{equation}
    \text{SDR@t} = \frac{\text{\# landmarks with MRE} \leq \text{t}}{\text{\# landmarks}} \times 100.
\end{equation}

\subsection{Datasets}
We evaluate five public and one private dataset spanning various 3D imaging modalities and anatomical regions. All test splits were used as hold-out test data.
\\
The \textbf{Mandibular Molar Landmarking (MML)} dataset \cite{he2024anchor} provides 648 CT images along with annotations of 14 dental landmarks targeting the crowns and roots of the second and third mandibular molars.
The dataset comes with the challenge of missing landmarks in cases where teeth are absent, structurally damaged, or have irregular root anatomy. However, we focus on predicting complete landmark annotations and use a subset that only included fully annotated cases, hereafter referred to as the complete MML subset.
In accordance with the official split, the complete subset contains 283 training, 56 validation, and 60 test cases. 
We used the official train/test split; the validation split was not used in this study.
\\
The \textbf{Anatomical Fiducials (AFIDs)} dataset \cite{taha2023magnetic,abbass2022application} consists of 132 T1 brain MRI images with 32 annotated brain landmarks as anatomical fiducials.
AFIDs is a collection of 4 subsets: 
(1) the AFIDs-HCP30 dataset (n=30), 3T scans from the Human Connectome Project (HCP) (\url{https://ida.loni.usc.edu/login.jsp}) \cite{van2012human};
(2) the AFIDs-OASIS30 dataset (n=30), 3T scans from the Open Access Series of Imaging Studies OASIS-1 \cite{marcus2007open}; 
(3) the London Health Sciences Center Parkinson’s disease (LHSCPD) dataset (n=40) containing gadolinium-enhanced images from a 1.5 T scanner \cite{ds004471:1.0.1}, and 
(4) the Stereotactic Neurosurgery (SNSX) \cite{ds004470:1.0.1} dataset (n=32) acquired with a 7T head-only scanner. 
Thus, this dataset is highly heterogeneous, with subdatasets differing in origin and imaging protocols. 
The human error on this dataset is reported as \SI{0.99}{\mm} with an inter-rater variability of \SI{1.48}{\mm}. 
As there is no official split, we performed a random split stratified across the four subsets into 110 training and 22 test cases, which will be published with the code. 
\\
The \textbf{Public Domain Database for Computational Anatomy (PDDCA)} dataset (version 1.4.1) \cite{Raudaschl2017PDDCA} contains CT scans of 33 patients with annotations for 5 bony landmarks in the head and neck area. We randomly split the data into 26 train and 7 test images, challenging the method's robustness in low data scenarios.
\\
The \textbf{Fetal pose estimation} dataset \cite{Chen2020fetuspose,Chen2024FetusMapV2} is a private dataset provided by Shenzhen University. It encompasses 1000 fetal ultrasound (US) images with 22 landmarks throughout the head, trunk and limbs, allowing to monitor fetal position and development. We used a split provided by the dataset authors, resulting in 800 training and 200 test cases.
\\
The \textbf{Fetal Tissue Annotation Challenge 2022 (FeTA22)} dataset contains super-resolution reconstructed T2 weighted MRIs of human fetal brains \cite{Payette2021FeTA21, Sanchez2024FeTA_Biometry}. Part of the data is provided by the University Children’s Hospital Zurich (Kispi) and is publicly available for research. This subset includes 68 cases with annotations for 10 landmarks, i.e. control points for 5 biometric measurements of the skull, brain and cerebellum. As the official test set is not public, we performed a custom random split into 65 training and 15 test cases.
\\
The \textbf{Fetal Cerebellum Landmark Detection (LFC)} dataset \cite{Gong2025DMGLD,Gong2025MethodBenchmark,Gong2025nnmamba} contains fetal brain T2 MRIs. The annotations target 12 landmarks, which represent control points for 6 biometric measurements concerning the skull, brain and cerebellum diameters, similar to FeTA22. We use the official train/test split of 120/60. For LFC, we evaluated both landmark detection and the resulting measurements, as the dataset primarily targets the downstream task of fetal brain biometry rather than precise landmark placement and some measurements can be taken in slightly shifted locations while still producing accurate biometry values.

\subsection{Benchmarking and Training Details} 
For nnLandmark, preprocessing and hyperparameters, such as patch size, batch size, network topology, are configured automatically by the framework based on dataset-specific properties. All nnLandmark experiments were performed using the \textit{3d\_fullres} preset and 5-fold cross-validation. We trained with the plain U-Net architecture, as well as variations with a ResNet-based encoder in two sizes, M and L (ResEncM/L), following official nnU-Net recommendations~\cite{isensee2024nnu}.
We compared nnLandmark to three recently published and state-of-the-art methods and toolkits. All compared methods were trained utilizing the respective official implementations and recommendations. The goal is to compare entire frameworks and repositories against each other to evaluate their generalizability to new datasets, without requiring any custom changes or elaborate hyperparameter tuning.
We additionally integrated the H3DE architecture \cite{huang2025H3DE} into nnLandmark to demonstrate its utility as a powerful, standardized method development framework.
\\
The \textbf{Hybrid-3D Network (H3DE-Net)} \cite{huang2025H3DE} integrates transformer-based attention modules within a U-Net-like CNN structure to effectively handle local feature extraction as well as global context modeling. 
The design of downsampling layers and window configuration restricts the input shape to be divisible by $64 \times 64 \times 32$. MML training was done using a random cropping data augmentation to $128 \times 128 \times 64$ voxels, to also fit the test image shape. For the remaining experiments, images were resized to $128^3$ voxels. 
\textbf{Landmarker} \cite{Jonkers2025landmarker, Jonkers2025uq} is a toolkit which offers useful modules for handling landmark data and frequently used architectures in the domain. They provide a set of default configurations that show promising results on the MML data \cite{Jonkers2025uq}, using a Flexible U-Net with EfficientNet backbone.
Following the practices of the authors, for MML, the training data were cropped based on the annotations to $128 \times 128 \times 64$ voxels, to fit the field of view in the test data. For the remaining datasets, images were resized to $128^3$. The models were trained as an ensemble using five different seeds. 
The \textbf{Super-Resolution U-Net (SR-UNet)} \cite{zhang2024SRLD} adopts pyramid pooling and super-resolution blocks to better preserve details and mitigate the error caused by downsampling and upsampling operations during training. 
All data was preprocessed using the published heatmap conversion script, which also includes resizing to $128^3$. For MML they report cropping the train data to $128 \times 128 \times 64$ voxels, so we use the same label-based crops as for Landmarker.

\begin{table}[t]
\centering
\floatconts
    {tab:res1}%
    {\caption{Results of landmark localisation accuracy of nnLandmark compared to current state-of-the-art methods, evaluated by MRE and micro standard deviation (std) on the respective on hold-out test data \sout{splits}. 
    All models were trained using the official code.}}%
    {\setlength{\tabcolsep}{3pt}}%
    \begin{tabular}{l c c c c}
    \toprule
    \textbf{Method} & \multicolumn{4}{c}{\textbf{MRE$\pm$Std [mm]}} \\
    & \textbf{MML} & \textbf{AFIDs} & \textbf{Fetal pose} & \textbf{PDDCA} \\
    \midrule
         H3DE & 1.81$\pm$1.15 & 4.28$\pm$2.09 & 6.07$\pm$6.44 & 8.21$\pm$4.62 \\
         SR-UNet & 10.01$\pm$10.37 & 3.37$\pm$1.97 & 5.93$\pm$6.09 & 7.74$\pm$4.45 \\
         landmarker & 10.58$\pm$13.92 & 2.86$\pm$4.12 & 5.37$\pm$7.99 & 4.98$\pm$2.71 \\
         \midrule
         nnLandmark H3DE & 1.63$\pm$1.16 & 1.79$\pm$1.05 & 4.25$\pm$6.33 & 3.31$\pm$2.31 \\
         \midrule
         nnLandmark & \underline{1.39$\pm$0.85} & \underline{1.55$\pm$1.01}  & 3.15$\pm$5.01 &  \textbf{2.51$\pm$2.53}\\
         nnLandmark ResEncM & \textbf{1.36$\pm$0.88} & \textbf{1.46$\pm$1.01} & \underline{3.06$\pm$4.51} & 2.82$\pm$3.27 \\
         nnLandmark ResEncL & 1.56$\pm$1.22 & 1.61$\pm$1.06 & \textbf{3.05$\pm$4.52} & \underline{2.72$\pm$2.76} \\
    \bottomrule
\end{tabular}
\end{table}

\begin{table}[t]
\centering
\floatconts
    {tab:res2}%
    {\caption{Results of the landmark localisation accuracy and resulting biometry measurements on the FeTA and LFC datasets.}}%
    {\setlength{\tabcolsep}{3pt}}%
    \begin{tabular}{l c c c c }
    \toprule
    \textbf{Method} & \multicolumn{2}{c}{\textbf{MRE $\pm$ Std [mm]}} & \multicolumn{2}{c}{\textbf{Biometry error $\pm$ Std [mm]}} \\
    & \textbf{FeTA} & \textbf{LFC} & \textbf{FeTA} & \textbf{LFC} \\
    \midrule
        H3DE & 3.61$\pm$2.74 & 4.22$\pm$4.21 & 3.33$\pm$2.84 & 2.85$\pm$2.32 \\
        SR-UNet & 3.41$\pm$2.73 & 3.92$\pm$3.62 & 3.61$\pm$2.70 & 5.92$\pm$5.55 \\
        landmarker & 3.26$\pm$3.76 & 4.02$\pm$5.13 & 2.15$\pm$3.28 & 1.78$\pm$4.25 \\
        \midrule
        nnLandmark H3DE & \textbf{2.71$\pm$3.09} & \textbf{3.72$\pm$4.47} & 2.12$\pm$1.85 & 1.52$\pm$1.24 \\
        \midrule
        nnLandmark & \underline{2.87$\pm$3.19} & 3.82$\pm$4.67 & \textbf{2.01$\pm$0.71} & 1.79$\pm$1.94 \\
        nnLandmark ResEncM & 4.03$\pm$10.14 & \underline{3.75$\pm$4.77} & \underline{2.03$\pm$2.62} & \underline{1.20$\pm$0.93} \\
        \textbf{nnLandmark ResEncL} & 3.67$\pm$8.13 & \underline{3.75$\pm$4.75} & 2.41$\pm$4.32 & \textbf{1.17$\pm$0.94} \\
    \bottomrule
\end{tabular}
\end{table}


\section{Results} 
nnLandmark, particularly in the ResEncM configuration, demonstrated the overall highest performance.
For the MML dataset, reproducibility results are reported in Appendix \ref{sec:app3_reproducibility}. For H3DE we only saw slight deviation from the reported results, attributable to some randomness during training \cite{huang2025H3DE}. For SR-UNet and Landmarker however results could  not be reproduced, with both yielding an MRE above \SI{10}{mm}, despite using the official repositories. 
This might be due to differences in handling the shift in field of view from whole-head CT during training to already cropped images in the test set. nnLandmark handles this inherently due to its patch-wise training scheme and sliding window prediction, and H3DE added random cropping during training. Landmarker required cropping the training images based on the labels to resemble the test shape. SR-UNet similarly reports random cropping of the images. Further, the reported Landmarker results were obtained on a custom randomized data split after cropping \cite{Jonkers2025uq}, hindering comparability. 
On AFIDs, all baselines showed moderate MRE of \SIrange{3}{4}{mm}. nnLandmark ResEncM achieved an MRE of \SI{1.46}{mm}, falling within the reported inter-rater variability of \SI{1.48}{mm} \cite{taha2023magnetic}.
For the fetal pose estimation task, all compared methods showed a moderate MRE of \SIrange{5}{6}{mm}. All models show a high standard deviation on this dataset, reflecting the challenging nature of the task, where the fetus can have varying positions in the uterus, arms and legs can be crossed and the respective landmarks consequently easily confused. 
The PDDCA results reveal that nnLandmark is the most robust in extreme low data scenarios, showing an MRE of $>$ \SI{3}{mm}.
For fetal brain annotation on the FeTA and LFC datasets, all models show a moderate MRE of about \SIrange{3}{4}{mm} and comparably high standard deviation, which could be attributed to the shifted annotation of some control points. A large error in landmark localization can still lead to accurate biometry measurements, which is also reflected in our results.\\
Integrating the H3DE architecture in our nnLandmark framework strongly improved accuracy compared to using H3DE with the published repository on all datasets. While it even achieved a slight advantage in FeTA and LFC landmark localization compared to nnLandmark U-Nets, this didn't show in the biometry measurement results. 
For all datasets, MRE for each landmark class individually are presented in Appendix~\ref{sec:app2_individual_mre}.
Randomly chosen example predictions for the nnLandmark ResEncM model for all datasets are shown in Figure~\ref{fig3}. Visual examples for all methods are given in the Appendix~\ref{sec:app4_qualitative}.

\begin{figure}[t]
    \centering
    \includegraphics[width=430pt]{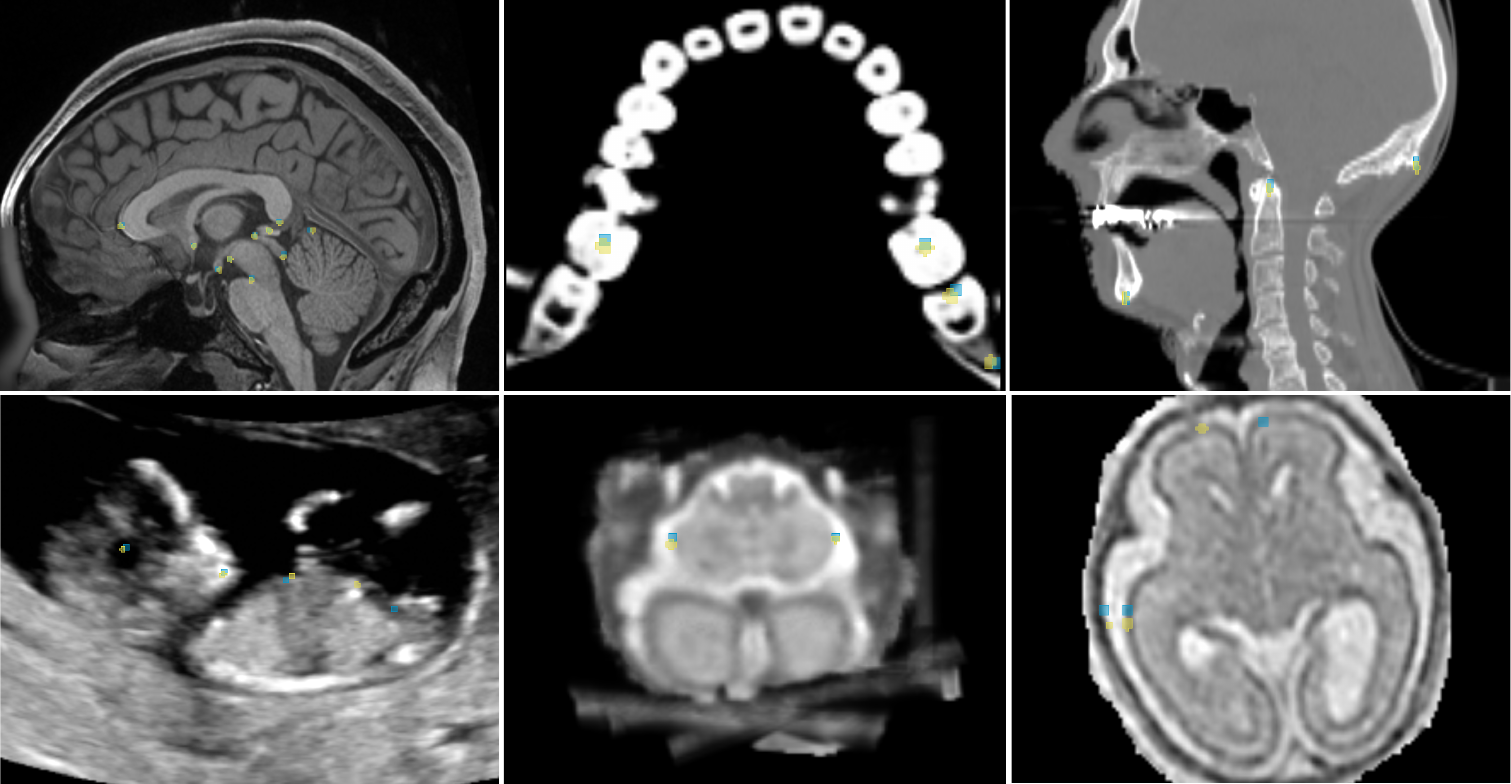}
    \caption{Visual examples for each dataset with ground truth (blue), nnLandmark ResEncM prediction (yellow). The ground truth is represented by a $3 \times 3 \times 3$ voxel segmentation. For visualization purposes, the predicted segmentation is generated by taking the 27 voxels with the highest values from each channel in the heatmap.}
    \label{fig3}
\end{figure}

\section{Discussion}
Current research in 3D landmark detection lacks the foundation needed for systematic progress, including transparent benchmarking, consistent baselines, and methods that reliably generalize across datasets (Figure~\ref{fig1}). 
Consequently, new methods are often not evaluated in a broader, standardized context and their translation to new datasets can require substantial manual effort, leading to a gap of more general solutions and insights. 
Tackling these pitfalls, we introduce nnLandmark, the first self-configuring framework for 3D medical landmark detection. 
Leveraging the established infrastructure of nnU-Net we inherit extensively optimized components for preprocessing, data augmentation and training while extending the framework with a dedicated heatmap representation, adapted loss computation, and coordinate prediction logic. 
This combination creates the first solution in the field to automatically adapt to new datasets without the need for expert intervention.
Thereby, nnLandmark occupies the unique position of serving as an out-of-the-box usable baseline as well as a flexible framework for method development and standardized evaluation, enabling transparent and comparable experimentation in the field.
\\
To ensure a comprehensive evaluation, we assessed nnLandmark five public and one private dataset spanning different modalities and anatomical regions and compare against three recently published methods and frameworks, H3DE, SR-UNet and Landmarker \cite{huang2025H3DE, zhang2024SRLD, Jonkers2025landmarker}. Although all three report strong performance on MML, systematic evaluation beyond this dataset has been missing. Our benchmarking closes this gap and highlights the need for broader evaluation as a standard practice in medical landmark detection.
The scarcity of established benchmarks and the common reliance on single, often private datasets have limited the comparability of methods. Our results show that performance shifts considerably across datasets, underlining the importance of designing and validating methods with generalization in mind. To facilitate broader adoption of multi-dataset evaluation, we release data conversion scripts for relevant public benchmarks, enabling straightforward use within the nnLandmark framework. \\
On the MML dataset, we also observed substantial variability among 3D U-Net baselines reported in the literature, with published MRE values ranging from \SIrange{1.90}{2.70}{mm} despite nominally identical architectures \cite{he2024anchor, huang2025H3DE, zhang2024SRLD}. These discrepancies demonstrate that architectural design alone is insufficient; optimal preprocessing, hyperparameter configuration, and training practices are equally important to achieve reliable performance.
nnLandmark addresses these issues as its self-configuring design eliminates the need for manual adjustments, providing a standardized, high-performing baseline for 3D landmark detection.
nnLandmark’s automatic adaptation to the dataset at hand further makes it the first framework to allow training state-of-the-art landmark detection models on new datasets without the need for expert knowledge and manual tuning.\\
Finally, the framework provides a controlled environment for developing and ablating new methodological ideas, relieving researchers from the need to build custom experimentation code while substantially improving the comparability of results. We illustrate this by integrating the H3DE architecture into nnLandmark, which further achieved improved performance compared to the official repository. These findings highlight the importance of a well-configured, standardized experimental environment for drawing meaningful, broadly applicable conclusions and reliably assessing methodological progress \cite{isensee2024nnu}.
\\
While nnLandmark addresses several long-standing challenges in landmark detection, some limitations remain. A limitation of storing labels as multi-label segmentation maps is the handling of closely spaced landmarks. Since each landmark is encoded as a $3 \times 3 \times 3$ voxel cube, two landmarks must be separated by at least three voxels to avoid overlap that would distort their encoded locations. In addition, nnLandmark’s current inference design always predicts a complete set of landmarks by taking the argmax of each heatmap channel. This does not account for anatomically absent landmarks, which occur, for example, in the full MML dataset where teeth may be missing. Handling presence or absence could be incorporated by estimating a confidence threshold from cross-validation and suppressing predictions below that threshold. The same mechanism could extend the framework to small object detection, where thresholding would allow multiple instances per class. Implementing reliable presence detection and multi-instance prediction is left for future work.

\section{Conclusion}
nnLandmark is introduced as a self-configuring deep learning framework for 3D medical landmark localization based on heatmap regression. 
It addresses three key pitfalls of the current literature: the lack of public benchmarking, inconsistent baseline implementations, and limited out-of-the-box usability. By conducting a benchmarking study across five public and one private dataset, nnLandmark establishes a transparent reference for evaluating existing and future methods.
Building on established components from nnU-Net, the framework translates self-configuration concepts to landmark detection, while providing a tailored heatmap generation, loss design, and inference logic. This enables robust generalization to new datasets without task-specific hyperparameter tuning or expert intervention. At the same time, nnLandmark offers a standardized, ready-to-use baseline and a flexible environment for method development, supporting reproducible experiments and systematic ablations. Together, these properties lay the groundwork for more rigorous and comparable research in 3D medical landmark detection, where novel ideas can be evaluated transparently and genuine methodological progress becomes measurable.

\clearpage  
\midlacknowledgments{
Regarding the AFIDs-HCP dataset: Data collection and sharing for this project was provided by the Human Connectome Project (HCP; Principal Investigators: Bruce Rosen, M.D., Ph.D., Arthur W. Toga, Ph.D., Van J. Weeden, MD). HCP funding was provided by the National Institute of Dental and Craniofacial Research (NIDCR), the National Institute of Mental Health (NIMH), and the National Institute of Neurological Disorders and Stroke (NINDS). HCP data are disseminated by the Laboratory of Neuro Imaging at the University of Southern California.
Regarding the AFIDs-OASIS dataset: Data were provided by
OASIS-1: Cross-Sectional: Principal Investigators: D. Marcus, R, Buckner, J, Csernansky J. Morris; P50 AG05681, P01 AG03991, P01 AG026276, R01 AG021910, P20 MH071616, U24 RR021382. Part of this work was funded by Helmholtz Imaging (HI), a platform of the Helmholtz Incubator on Information and Data Science.}

\bibliography{references}

@article{isensee2021nnu,
  title={nnU-Net: a self-configuring method for deep learning-based biomedical image segmentation},
  author={Isensee, Fabian and Jaeger, Paul F and Kohl, Simon AA and Petersen, Jens and Maier-Hein, Klaus H},
  journal={Nature methods},
  volume={18},
  number={2},
  pages={203--211},
  year={2021},
  publisher={Nature Publishing Group}
}

@inproceedings{isensee2024nnu,
  title={nnu-net revisited: A call for rigorous validation in 3d medical image segmentation},
  author={Isensee, Fabian and Wald, Tassilo and Ulrich, Constantin and Baumgartner, Michael and Roy, Saikat and Maier-Hein, Klaus and Jaeger, Paul F},
  booktitle={International Conference on Medical Image Computing and Computer-Assisted Intervention},
  pages={488--498},
  year={2024},
  organization={Springer}
}

@misc{ronneberger2015unet,
      title={U-Net: Convolutional Networks for Biomedical Image Segmentation}, 
      author={Olaf Ronneberger and Philipp Fischer and Thomas Brox},
      year={2015},
      eprint={1505.04597},
      archivePrefix={arXiv},
      primaryClass={cs.CV},
      url={https://arxiv.org/abs/1505.04597}, 
}

@inproceedings{cciccek20163dunet,
  title={3D U-Net: learning dense volumetric segmentation from sparse annotation},
  author={{\c{C}}i{\c{c}}ek, {\"O}zg{\"u}n and Abdulkadir, Ahmed and Lienkamp, Soeren S and Brox, Thomas and Ronneberger, Olaf},
  booktitle={Medical Image Computing and Computer-Assisted Intervention--MICCAI 2016: 19th International Conference, Athens, Greece, October 17-21, 2016, Proceedings, Part II 19},
  pages={424--432},
  year={2016},
  organization={Springer}
}

@misc{huang2025H3DE,
      title={H3DE-Net: Efficient and Accurate 3D Landmark Detection in Medical Imaging}, 
      author={Zhen Huang and Ronghao Xu and Xiaoqian Zhou and Yangbo Wei and Suhua Wang and Xiaoxin Sun and Han Li and Qingsong Yao},
      year={2025},
      eprint={2502.14221},
      archivePrefix={arXiv},
      primaryClass={cs.CV},
      url={https://arxiv.org/abs/2502.14221}, 
}

@article{zhang2024SRLD,
  title={Super-resolution landmark detection networks for medical images},
  author={Zhang, Runshi and Mo, Hao and Hu, Weini and Jie, Bimeng and Xu, Lin and He, Yang and Ke, Jia and Wang, Junchen},
  journal={Computers in Biology and Medicine},
  volume={182},
  pages={109095},
  year={2024},
  publisher={Elsevier}
}

@article{Jonkers2025landmarker,
    title = {landmarker: A Toolkit for Anatomical Landmark Localization in 2D/3D Images},
    journal = {SoftwareX},
    volume = {30},
    pages = {102165},
    year = {2025},
    issn = {2352-7110},
    doi = {10.1016/j.softx.2025.102165},
    author = {Jonkers, Jef and Duchateau, Luc and Van Wallendael, Glenn and Van Hoecke, Sofie}
}

@misc{Jonkers2025uq,
    author = {Jonkers, Jef and Coopman, Frank and Duchateau, Luc and Van Wallendael, Glenn and Hoecke, Sofie},
    year = {2025},
    month = {03},
    title = {Reliable uncertainty quantification for 2D/3D anatomical landmark localization using multi-output conformal prediction},
    archivePrefix={arXiv},
    doi = {10.48550/arXiv.2503.14106}
}

@article{he2024anchor,
  title={Anchor Ball Regression Model for large-scale 3D skull landmark detection},
  author={He, Tao and Xu, Guikun and Cui, Li and Tang, Wei and Long, Jie and Guo, Jixiang},
  journal={Neurocomputing},
  volume={567},
  pages={127051},
  year={2024},
  publisher={Elsevier}
}

@article{taha2023magnetic,
  title={Magnetic resonance imaging datasets with anatomical fiducials for quality control and registration},
  author={Taha, Alaa and Gilmore, Greydon and Abbass, Mohamad and Kai, Jason and Kuehn, Tristan and Demarco, John and Gupta, Geetika and Zajner, Chris and Cao, Daniel and Chevalier, Ryan and others},
  journal={Scientific Data},
  volume={10},
  number={1},
  pages={449},
  year={2023},
  publisher={Nature Publishing Group UK London}
}

@article{abbass2022application,
  title={Application of the anatomical fiducials framework to a clinical dataset of patients with Parkinson’s disease},
  author={Abbass, Mohamad and Gilmore, Greydon and Taha, Alaa and Chevalier, Ryan and Jach, Magdalena and Peters, Terry M and Khan, Ali R and Lau, Jonathan C},
  journal={Brain Structure and Function},
  pages={1--13},
  year={2022},
  publisher={Springer}
}

@article{marcus2007open,
  title={Open Access Series of Imaging Studies (OASIS): cross-sectional MRI data in young, middle aged, nondemented, and demented older adults},
  author={Marcus, Daniel S and Wang, Tracy H and Parker, Jamie and Csernansky, John G and Morris, John C and Buckner, Randy L},
  journal={Journal of cognitive neuroscience},
  volume={19},
  number={9},
  pages={1498--1507},
  year={2007},
  publisher={MIT Press One Rogers Street, Cambridge, MA 02142-1209, USA journals-info~…}
}

@dataset{ds004471:1.0.1,
  author = {Mohamad Abbass and Greydon Gilmore and Alaa Taha and Ryan Chevalier and Magdalena Jach and Terry M. Peters and Ali R. Khan and Jonathan C. Lau},
  title = {"London Heath Sciences Center Parkinson's Disease Dataset (LHSCPD)"},
  year = {2023},
  doi = {doi:10.18112/openneuro.ds004471.v1.0.1},
  publisher = {OpenNeuro}
}

@dataset{ds004470:1.0.1,
  author = {Jonathan C. Lau and Yiming Xiao and Roy A. M. Haast and Greydon Gilmore and Kâmil Uludağ and Keith W. MacDougall and Ravi S. Menon and Andrew G. Parrent and Terry M. Peters and Ali R. Khan},
  title = {"Stereotactic Neurosurgery Dataset (SNSX)"},
  year = {2023},
  doi = {doi:10.18112/openneuro.ds004470.v1.0.1},
  publisher = {OpenNeuro}
}

@article{van2012human,
  title={The Human Connectome Project: a data acquisition perspective},
  author={Van Essen, David C and Ugurbil, Kamil and Auerbach, Edward and Barch, Deanna and Behrens, Timothy EJ and Bucholz, Richard and Chang, Acer and Chen, Liyong and Corbetta, Maurizio and Curtiss, Sandra W and others},
  journal={Neuroimage},
  volume={62},
  number={4},
  pages={2222--2231},
  year={2012},
  publisher={Elsevier}
}

@article{Gong2025DMGLD, title={Domain Generalized Medical Landmark Detection via Robust Boundary-Aware Pre-Training}, volume={39}, url={https://ojs.aaai.org/index.php/AAAI/article/view/32323}, DOI={10.1609/aaai.v39i3.32323}, number={3}, journal={Proceedings of the AAAI Conference on Artificial Intelligence}, author={Gong, Haifan and Lu, Yu and Wan, Xiang and Li, Haofeng}, year={2025}, month={Apr.}, pages={3140-3148} }

@ARTICLE{Gong2025MethodBenchmark,
  author={Gong, Haifan and Liu, Huixian and Wang, Yitao and Liu, Xiaoling and Wan, Xiang and Shi, Qiao and Li, Haofeng},
  journal={IEEE Journal of Biomedical and Health Informatics}, 
  title={Fetal Cerebellum Landmark Detection Based on 3D MRI: Method and Benchmark}, 
  year={2025},
  volume={29},
  number={8},
  pages={5712-5721},
  doi={10.1109/JBHI.2025.3559702}}

@INPROCEEDINGS{Gong2025nnmamba,
  author={Gong, Haifan and Kang, Luoyao and Wang, Yitao and Wang, Yihan and Wan, Xiang and Wu, Xusheng and Li, Haofeng},
  booktitle={2025 IEEE 22nd International Symposium on Biomedical Imaging (ISBI)}, 
  title={Nnmamba: 3D Biomedical Image Segmentation, Classification and Landmark Detection with State Space Model}, 
  year={2025},
  volume={},
  number={},
  pages={1-5},
  doi={10.1109/ISBI60581.2025.10980694}}

@INPROCEEDINGS{Chen2020fetuspose,
  author={Chen, Chaoyu and Yang, Xin and Huang, Ruobing and Shi, Wenlong and Liu, Shengfeng and Lin, Mingrong and Huang, Yuhao and Yang, Yong and Zhang, Yuanji and Luo, Huanjia and Huang, Yankai and Xiong, Yi and Ni, Dong},
  booktitle={2020 IEEE 17th International Symposium on Biomedical Imaging (ISBI)}, 
  title={Region Proposal Network with Graph Prior and Iou-Balance Loss for Landmark Detection in 3D Ultrasound}, 
  year={2020},
  volume={},
  number={},
  pages={1-5},
  doi={10.1109/ISBI45749.2020.9098368}}

@article{Chen2024FetusMapV2,
title = {FetusMapV2: Enhanced fetal pose estimation in 3D ultrasound},
journal = {Medical Image Analysis},
volume = {91},
pages = {103013},
year = {2024},
issn = {1361-8415},
doi = {https://doi.org/10.1016/j.media.2023.103013},
url = {https://www.sciencedirect.com/science/article/pii/S1361841523002736},
author = {Chaoyu Chen and Xin Yang and Yuhao Huang and Wenlong Shi and Yan Cao and Mingyuan Luo and Xindi Hu and Lei Zhu and Lequan Yu and Kejuan Yue and Yuanji Zhang and Yi Xiong and Dong Ni and Weijun Huang},
}

@article{Raudaschl2017PDDCA,
author = {Raudaschl, Patrik F. and Zaffino, Paolo and Sharp, Gregory C. and Spadea, Maria Francesca and Chen, Antong and Dawant, Benoit M. and Albrecht, Thomas and Gass, Tobias and Langguth, Christoph and Lüthi, Marcel and Jung, Florian and Knapp, Oliver and Wesarg, Stefan and Mannion-Haworth, Richard and Bowes, Mike and Ashman, Annaliese and Guillard, Gwenael and Brett, Alan and Vincent, Graham and Orbes-Arteaga, Mauricio and Cárdenas-Peña, David and Castellanos-Dominguez, German and Aghdasi, Nava and Li, Yangming and Berens, Angelique and Moe, Kris and Hannaford, Blake and Schubert, Rainer and Fritscher, Karl D.},
title = {Evaluation of segmentation methods on head and neck CT: Auto-segmentation challenge 2015},
journal = {Medical Physics},
volume = {44},
number = {5},
pages = {2020-2036},
doi = {https://doi.org/10.1002/mp.12197},
year = {2017}
}

@article{Payette2021FeTA21,
  author       = {Payette, Kelly and de Dumast, Priscille and Kebiri, Hamza and Ezhov, Ivan and Paetzold, Johannes C. and Shit, Suprosanna and Iqbal, Asim and Khan, Romesa and Kottke, Raimund and Grehten, Patrice and Ji, Hui and Lanczi, Levente and Nagy, Marianna and Beresova, Monika and Nguyen, Thi Dao and Natalucci, Giancarlo and Karayannis, Theofanis and Menze, Bjoern and Bach Cuadra, Meritxell and Jakab, Andras},
  title        = {An automatic multi-tissue human fetal brain segmentation benchmark using the Fetal Tissue Annotation Dataset},
  journaltitle  = {Scientific Data},
  date         = {2021-07-06},
  volume       = {8},
  number       = {1},
  pages        = {167},
  doi          = {10.1038/s41597-021-00946-3},
  url          = {https://doi.org/10.1038/s41597-021-00946-3},
  issn         = {2052-4463},
}

@misc{Sanchez2024FeTA_Biometry,
  author       = {Sanchez, Thomas and
                  Gomez, Yvan and
                  Licandro, Roxane and
                  Payette, Kelly and
                  Jakab, Andras and
                  Koob, Meriam and
                  Bach Cuadra, Meritxell},
  title        = {Fetal Tissue Annotation Challenge (FeTA) Biometry
                   - MICCAI 2024
                  },
  month        = may,
  year         = 2024,
  publisher    = {Zenodo},
  version      = 1,
  doi          = {10.5281/zenodo.11192452},
  url          = {https://doi.org/10.5281/zenodo.11192452},
}

@article{schwendicke2021deep,
  title={Deep learning for cephalometric landmark detection: systematic review and meta-analysis},
  author={Schwendicke, Falk and Chaurasia, Akhilanand and Arsiwala, Lubaina and Lee, Jae-Hong and Elhennawy, Karim and Jost-Brinkmann, Paul-Georg and Demarco, Flavio and Krois, Joachim},
  journal={Clinical oral investigations},
  volume={25},
  number={7},
  pages={4299--4309},
  year={2021},
  publisher={Springer}
}

@article{serafin2023accuracy,
  title={Accuracy of automated 3D cephalometric landmarks by deep learning algorithms: systematic review and meta-analysis},
  author={Serafin, Marco and Baldini, Benedetta and Cabitza, Federico and Carrafiello, Gianpaolo and Baselli, Giuseppe and Del Fabbro, Massimo and Sforza, Chiarella and Caprioglio, Alberto and Tartaglia, Gianluca M},
  journal={La radiologia medica},
  volume={128},
  number={5},
  pages={544--555},
  year={2023},
  publisher={Springer}
}

@article{singh20203d,
  title={3D deep learning on medical images: a review},
  author={Singh, Satya P and Wang, Lipo and Gupta, Sukrit and Goli, Haveesh and Padmanabhan, Parasuraman and Guly{\'a}s, Bal{\'a}zs},
  journal={Sensors},
  volume={20},
  number={18},
  pages={5097},
  year={2020},
  publisher={MDPI}
}

@article{chen2021fast,
  title={Fast and accurate craniomaxillofacial landmark detection via 3D faster R-CNN},
  author={Chen, Xiaoyang and Lian, Chunfeng and Deng, Hannah H and Kuang, Tianshu and Lin, Hung-Ying and Xiao, Deqiang and Gateno, Jaime and Shen, Dinggang and Xia, James J and Yap, Pew-Thian},
  journal={IEEE transactions on medical imaging},
  volume={40},
  number={12},
  pages={3867--3878},
  year={2021},
  publisher={IEEE}
}

@article{chen2022structure,
  title={Structure-aware long short-term memory network for 3D cephalometric landmark detection},
  author={Chen, Runnan and Ma, Yuexin and Chen, Nenglun and Liu, Lingjie and Cui, Zhiming and Lin, Yanhong and Wang, Wenping},
  journal={IEEE Transactions on Medical Imaging},
  volume={41},
  number={7},
  pages={1791--1801},
  year={2022},
  publisher={IEEE}
}

@inproceedings{pfister2015flowing,
  title={Flowing convnets for human pose estimation in videos},
  author={Pfister, Tomas and Charles, James and Zisserman, Andrew},
  booktitle={Proceedings of the IEEE international conference on computer vision},
  pages={1913--1921},
  year={2015}
}

@inproceedings{payer2016regressing,
  title={Regressing heatmaps for multiple landmark localization using CNNs},
  author={Payer, Christian and {\v{S}}tern, Darko and Bischof, Horst and Urschler, Martin},
  booktitle={International conference on medical image computing and computer-assisted intervention},
  pages={230--238},
  year={2016},
  organization={Springer}
}

@article{cui2025pseudo,
  title={A pseudo-3D coarse-to-fine architecture for 3D medical landmark detection},
  author={Cui, Li and Liu, Boyan and Xu, Guikun and Guo, Jixiang and Tang, Wei and He, Tao},
  journal={Neurocomputing},
  volume={614},
  pages={128782},
  year={2025},
  publisher={Elsevier}
}

@article{liu2023joint,
  title={Joint cranial bone labeling and landmark detection in pediatric CT images using context encoding},
  author={Liu, Jiawei and Xing, Fuyong and Shaikh, Abbas and French, Brooke and Linguraru, Marius George and Porras, Antonio R},
  journal={IEEE transactions on medical imaging},
  volume={42},
  number={10},
  pages={3117--3126},
  year={2023},
  publisher={IEEE}
}

@InProceedings{pang2024prior,
  title = 	 {Prior Guided 3D Medical Image Landmark Localization},
  author =       {Pang, Yijie and Cheng, Pujin and Lyu, Junyan and Lin, Fan and Tang, Xiaoying},
  booktitle = 	 {Medical Imaging with Deep Learning},
  pages = 	 {1163--1175},
  year = 	 {2024},
  editor = 	 {Oguz, Ipek and Noble, Jack and Li, Xiaoxiao and Styner, Martin and Baumgartner, Christian and Rusu, Mirabela and Heinmann, Tobias and Kontos, Despina and Landman, Bennett and Dawant, Benoit},
  volume = 	 {227},
  series = 	 {Proceedings of Machine Learning Research},
  month = 	 {10--12 Jul},
  publisher =    {PMLR},
  url = 	 {https://proceedings.mlr.press/v227/pang24a.html},
}

@article{barough2025brainsignsnet,
  title={BrainSignsNET: Deep Learning-Based 3D Anatomical Landmark Detection in Human Brain Imaging},
  author={Barough, Siavash Shirzadeh and Ventura, Catalina and Bilgel, Murat and Albert, Marilyn S and Miller, Michael I and Moghekar, Abhay},
  journal={medRxiv},
  pages={2025--07},
  year={2025},
  publisher={Cold Spring Harbor Laboratory Press}
}

@article{kang2021CephReinforcement,
  title={3D cephalometric landmark detection by multiple stage deep reinforcement learning},
  author={Kang, Sung Ho and Jeon, Kiwan and Kang, Sang-Hoon and Lee, Sang-Hwy},
  journal={Scientific reports},
  volume={11},
  number={1},
  pages={17509},
  year={2021},
  publisher={Nature Publishing Group UK London}
}

@inproceedings{jiang2022cephalformer,
  title={Cephalformer: Incorporating global structure constraint into visual features for general cephalometric landmark detection},
  author={Jiang, Yankai and Li, Yiming and Wang, Xinyue and Tao, Yubo and Lin, Jun and Lin, Hai},
  booktitle={International conference on medical image computing and computer-assisted intervention},
  pages={227--237},
  year={2022},
  organization={Springer}
}

@inproceedings{shi2025EndTooth,
  title={End-to-End 3D Tooth Landmark Detection with Fuzzy Tooth Localization},
  author={Shi, Kaibo and Jin, Hairong and Zheng, Youyi},
  booktitle={International Conference on Medical Image Computing and Computer-Assisted Intervention},
  pages={170--180},
  year={2025},
  organization={Springer}
}

@article{dai2024NasalBone,
  title={Cone-beam CT landmark detection for measuring basal bone width: a retrospective validation study},
  author={Dai, Juan and Guo, Xinge and Zhang, Hongyuan and Xie, Haoyu and Huang, Jiahui and Huang, Qiangtai and Huang, Bingsheng},
  journal={BMC Oral Health},
  volume={24},
  number={1},
  pages={1091},
  year={2024},
  publisher={Springer}
}

@article{stebani2023InnerEar,
  title={Towards fully automated inner ear analysis with deep-learning-based joint segmentation and landmark detection framework},
  author={Stebani, Jannik and Blaimer, Martin and Zabler, Simon and Neun, Tilmann and Pelt, Dani{\"e}l M and Rak, Kristen},
  journal={Scientific Reports},
  volume={13},
  number={1},
  pages={19057},
  year={2023},
  publisher={Nature Publishing Group UK London}
}

@article{baksi2021SoftTissue,
  title={Accuracy of an automated method of 3D soft tissue landmark detection},
  author={Baksi, Sanjana and Freezer, Simon and Matsumoto, Takeshi and Dreyer, Craig},
  journal={European journal of orthodontics},
  volume={43},
  number={6},
  pages={622--630},
  year={2021},
  publisher={Oxford University Press UK}
}

@article{gillot2023cbct,
  title={Automatic landmark identification in cone-beam computed tomography},
  author={Gillot, Maxime and Miranda, Felicia and Baquero, Baptiste and Ruellas, Antonio and Gurgel, Marcela and Al Turkestani, Najla and Anchling, Luc and Hutin, Nathan and Biggs, Elizabeth and Yatabe, Marilia and others},
  journal={Orthodontics \& craniofacial research},
  volume={26},
  number={4},
  pages={560--567},
  year={2023},
  publisher={Wiley Online Library}
}

@inproceedings{salari2023uncertainty,
  title={Uncertainty-aware transformer model for anatomical landmark detection in paraspinal muscle MRIs},
  author={Salari, Soorena and Rasoulian, Amirhossein and Battie, Michele and Fortin, Maryse and Rivaz, Hassan and Xiao, Yiming},
  booktitle={Medical Imaging 2023: Image Processing},
  volume={12464},
  pages={246--252},
  year={2023},
  organization={SPIE}
}

@ARTICLE{wu2022MeshTooth,
  author={Wu, Tai-Hsien and Lian, Chunfeng and Lee, Sanghee and Pastewait, Matthew and Piers, Christian and Liu, Jie and Wang, Fan and Wang, Li and Chiu, Chiung-Ying and Wang, Wenchi and Jackson, Christina and Chao, Wei-Lun and Shen, Dinggang and Ko, Ching-Chang},
  journal={IEEE Transactions on Medical Imaging}, 
  title={Two-Stage Mesh Deep Learning for Automated Tooth Segmentation and Landmark Localization on 3D Intraoral Scans}, 
  year={2022},
  volume={41},
  number={11},
  pages={3158-3166},
  doi={10.1109/TMI.2022.3180343}}

@ARTICLE{Li2023SpatialKnee,
  author={Li, Xiang and Lv, Songcen and Li, Minglei and Zhang, Jiusi and Jiang, Yuchen and Qin, Yong and Luo, Hao and Yin, Shen},
  journal={IEEE Transactions on Medical Imaging}, 
  title={SDMT: Spatial Dependence Multi-Task Transformer Network for 3D Knee MRI Segmentation and Landmark Localization}, 
  year={2023},
  volume={42},
  number={8},
  pages={2274-2285},
  doi={10.1109/TMI.2023.3247543}}

@article{lu2023cmf,
  title={CMF-Net: craniomaxillofacial landmark localization on CBCT images using geometric constraint and transformer},
  author={Lu, Gang and Shu, Huazhong and Bao, Han and Kong, Youyong and Zhang, Chen and Yan, Bin and Zhang, Yuanxiu and Coatrieux, Jean-Louis},
  journal={Physics in Medicine \& Biology},
  volume={68},
  number={9},
  pages={095020},
  year={2023},
  publisher={IOP Publishing}
}

@ARTICLE{Lang2022CraniomaxillofacialMaskRCNN,
  author={Lang, Yankun and Lian, Chunfeng and Xiao, Deqiang and Deng, Hannah and Thung, Kim-Han and Yuan, Peng and Gateno, Jaime and Kuang, Tianshu and Alfi, David M. and Wang, Li and Shen, Dinggang and Xia, James J. and Yap, Pew-Thian},
  journal={IEEE Transactions on Medical Imaging}, 
  title={Localization of Craniomaxillofacial Landmarks on CBCT Images Using 3D Mask R-CNN and Local Dependency Learning}, 
  year={2022},
  volume={41},
  number={10},
  pages={2856-2866},
  doi={10.1109/TMI.2022.3174513}}

@inproceedings{lopez2021facial,
  title={Facial and cochlear nerves characterization using deep reinforcement learning for landmark detection},
  author={L{\'o}pez Diez, Paula and Sundgaard, Josefine Vilsb{\o}ll and Patou, Fran{\c{c}}ois and Margeta, Jan and Paulsen, Rasmus Reinhold},
  booktitle={International Conference on Medical Image Computing and Computer-Assisted Intervention},
  pages={519--528},
  year={2021},
  organization={Springer}
}

@article{zhu2022novel,
  title={A novel method for 3D knee anatomical landmark localization by combining global and local features},
  author={Zhu, Junjun and Zhao, Qijie and Zhu, Junhao and Zhou, Anwen and Shao, Hui},
  journal={Machine Vision and Applications},
  volume={33},
  number={4},
  pages={52},
  year={2022},
  publisher={Springer}
}

@inproceedings{lang2022dentalpointnet,
  title={DentalPointNet: landmark localization on high-resolution 3D digital dental models},
  author={Lang, Yankun and Chen, Xiaoyang and Deng, Hannah H and Kuang, Tianshu and Barber, Joshua C and Gateno, Jaime and Yap, Pew-Thian and Xia, James J},
  booktitle={International conference on medical image computing and computer-assisted intervention},
  pages={444--452},
  year={2022},
  organization={Springer}
}

@article{dot2022automatic,
  title={Automatic 3-dimensional cephalometric landmarking via deep learning},
  author={Dot, Gauthier and Schouman, Thomas and Chang, Shaole and Rafflenbeul, Fr{\'e}d{\'e}ric and Kerbrat, Adeline and Rouch, Philippe and Gajny, Laurent},
  journal={Journal of dental research},
  volume={101},
  number={11},
  pages={1380--1387},
  year={2022},
  publisher={SAGE Publications Sage CA: Los Angeles, CA}
}

@article{torosdagli2023relational,
  title={Relational reasoning network for anatomical landmarking},
  author={Torosdagli, Neslisah and Anwar, Syed and Verma, Payal and Liberton, Denise K and Lee, Janice S and Han, Wade W and Bagci, Ulas},
  journal={Journal of Medical Imaging},
  volume={10},
  number={2},
  pages={024002--024002},
  year={2023},
  publisher={Society of Photo-Optical Instrumentation Engineers}
}

\clearpage
\appendix

\section{List of analyzed publications for Figure 1}
\label{sec:app0_fig1}
We identified current pitfalls in the 3D medical landmark detection literature based on the following representative list of relevant methodological publications since 2021: \cite{huang2025H3DE, zhang2024SRLD, Jonkers2025landmarker, Gong2025nnmamba, he2024anchor, chen2021fast, Chen2024FetusMapV2, cui2025pseudo, liu2023joint, pang2024prior, Gong2025MethodBenchmark, barough2025brainsignsnet, chen2022structure, kang2021CephReinforcement, jiang2022cephalformer, shi2025EndTooth, dai2024NasalBone, stebani2023InnerEar, baksi2021SoftTissue, gillot2023cbct, salari2023uncertainty, wu2022MeshTooth, Li2023SpatialKnee, lu2023cmf, Lang2022CraniomaxillofacialMaskRCNN, lopez2021facial, zhu2022novel, lang2022dentalpointnet, dot2022automatic, torosdagli2023relational}

\section{Ablation Study on Hyperparameters}
\label{sec:app0_ablations}

\begin{table}[H]
\centering
\floatconts
    {tab:appendix_ablations}%
    {\caption{Ablation study assessing the robustness of our hyperparameter choices. EDT denotes the Euclidean distance transform radius used to generate the heatmap supervision for landmark localization during training. We additionally compare alternative loss functions, including MSE and BCE Topk with varying k. nnLandmark uses EDT 15 and BCE Top20 loss.}}
    {\setlength{\tabcolsep}{3pt}}%
    \begin{tabular}{l c c }
    \toprule
    \textbf{Parameter} & \multicolumn{2}{c}{\textbf{MRE [mm]}} \\
    & \textbf{Afids} & \textbf{Fetal Pose} \\
    \midrule
        EDT 7   & 1.65  & 17.59 \\
        EDT 11  & 1.59  & 3.21  \\
        EDT 15  & 1.66 & 2.87  \\
        EDT 19  & 1.64  & 2.85  \\
        EDT 23  & 1.68  & 2.82  \\
        \midrule
        MSE loss & 12.06 & 52.48 \\
        \midrule
        BCE Top 10  & 1.63  & 2.85  \\
        BCE Top 20  & 1.66 & 2.87  \\
        BCE Top 30  & 1.62  & 3.00  \\
        BCE Top 40  & 1.62  & 3.00  \\
        BCE Top 50  & 1.67  & 3.02  \\
        BCE Top 60  & 1.68  & 3.21  \\
        BCE Top 70  & 1.69  & 3.15  \\
        BCE Top 80  & 1.72  & 3.15  \\
        BCE Top 90  & 1.76  & 3.23  \\
        BCE Top 100 & 1.76  & 3.19  \\
    \bottomrule
\end{tabular}
\end{table}

\section{Extended Results Including Success Detection Rate (SDR)}
\label{sec:app1_sdr}

\begin{table}[H]
\centering
\floatconts
    {tab:appendix_sdr}%
    {\caption{Extended results table with SDR. 
    All results are done on the hold-out testsplits.}}%
    {\setlength{\tabcolsep}{3pt}}%
    \begin{tabular}{l c c c c}
    \toprule
    \textbf{Method} & \textbf{MRE$\pm$Std} & \multicolumn{3}{c}{\textbf{SDR [\%]}} \\
    & \textbf{[mm]} & \textbf{2 mm} & \textbf{3 mm} & \textbf{4 mm} \\
    \midrule
    \multicolumn{5}{l}{\textbf{AFIDs} \#samples 22, \#landmarks 32}\\
    \midrule
         H3DE \cite{huang2025H3DE} & 4.28$\pm$2.09 & 13.07 & 28.98 & 51.28 \\
         SR-UNet \cite{zhang2024SRLD} & 3.37$\pm$1.97 & 25.43 & 50.99 & 70.03 \\
         landmarker \cite{Jonkers2025landmarker} & 2.86$\pm$4.12 & 46.16 & 62.22 & 74.57 \\
         \midrule
         nnLandmark H3DE & 1.79$\pm$1.05 & 67.90 & 88.78 & 97.02 \\
         \midrule
         nnLandmark & 1.55$\pm$1.01 & 76.85 & 93.61 & 97.87 \\
         \textbf{nnLandmark ResEncM} & \textbf{1.46$\pm$1.01} & \textbf{81.82} & \textbf{94.74} & \textbf{98.15} \\
         nnLandmark ResEncL & 1.61$\pm$1.06 & 75.71 & 92.90 & 97.44 \\
    \midrule
    \multicolumn{5}{l}{\textbf{MML complete subset} \#samples 60, \#landmarks 14}\\
    \midrule
         H3DE \cite{huang2025H3DE} & 1.81$\pm$1.15 & 67.14 & 89.52 & 97.14 \\
         SR-UNet \cite{zhang2024SRLD} & 10.01$\pm$10.37 & 5.24 & 13.93 & 24.17 \\
         landmarker \cite{Jonkers2025landmarker} & 10.58$\pm$13.92 & 24.05 & 34.05 & 42.62 \\
         \midrule
         nnLandmark H3DE & 1.63$\pm$1.16 & 72.50 & 91.19 & 97.26 \\
         \midrule
         nnLandmark & 1.39$\pm$0.85 & 80.00 & 95.24 & 98.57 \\
         \textbf{nnLandmark ResEncM} & \textbf{1.36$\pm$0.88} & \textbf{82.02} & \textbf{95.48} & \textbf{98.69} \\\
         nnLandmark ResEncL & 1.59$\pm$1.22 & 75.24 & 92.50 & 97.98 \\
    \midrule
    \multicolumn{5}{l}{\textbf{Fetal pose} \#samples 200, \#landmarks 22}\\
    \midrule
         H3DE \cite{huang2025H3DE} & 6.07$\pm$6.44 & 11.21 & 27.81 & 45.02 \\
         SR-UNet \cite{zhang2024SRLD} & 66.11$\pm$19.15 & 0.04 & 0.04 & 0.04 \\
         landmarker \cite{Jonkers2025landmarker} & 5.37$\pm$7.99 & 36.45 & 56.16 & 67.36 \\
         \midrule
         nnLandmark H3DE & 4.35$\pm$6.33 & 44.77 & 64.23 & 74.77 \\
         \midrule
         nnLandmark & 3.15$\pm$5.01 & 49.77 & 70.86 & 82.16 \\
         nnLandmark ResEncM & 3.06$\pm$4.51 & 50.80 & 70.64 & 81.68\\
         \textbf{nnLandmark ResEncL} & \textbf{3.05$\pm$4.52} & \textbf{51.19} & \textbf{71.09} & \textbf{81.80} \\
    \midrule
        \multicolumn{5}{l}{\textbf{PDDCA} \#samples 7, \#landmarks 5}\\
    \midrule
        H3DE \cite{huang2025H3DE} & 8.21$\pm$4.62 & 2.86 & 5.71 & 11.43 \\
        SR-UNet \cite{zhang2024SRLD} & 7.74$\pm$4.45 & 2.86 & 11.43 & 28.57 \\
        landmarker \cite{Jonkers2025landmarker} & 4.98$\pm$2.71 & 2.86 & 11.43 & 37.14 \\
        \midrule
        nnLandmark H3DE & 3.31$\pm$2.31 & 25.71 & 54.29 & 77.14 \\
        \midrule
        nnLandmark & 2.51$\pm$2.53 & 45.71 & 74.29 & 91.43 \\
        \textbf{nnLandmark ResEncM} & \textbf{2.82$\pm$3.27} & \textbf{45.71} & \textbf{71.43} & \textbf{88.57} \\
        nnLandmark ResEncL & 2.72$\pm$2.76 & 40.00 & 68.57 & 88.57 \\
    \bottomrule
\end{tabular}
\end{table}

\begin{table}[H]
\centering
\floatconts
    {tab:appendix_sdr}%
    {\caption{Extended results table with SDR. 
    All results are done on the hold-out testsplits.}}%
    {\setlength{\tabcolsep}{3pt}}%
    \begin{tabular}{l c c c c}
    \toprule
    \textbf{Method} & \textbf{MRE$\pm$Std} & \multicolumn{3}{c}{\textbf{SDR [\%]}} \\
    & \textbf{[mm]} & \textbf{2 mm} & \textbf{3 mm} & \textbf{4 mm} \\
    \midrule
        \multicolumn{5}{l}{\textbf{FeTA} \#samples 15, \#landmarks 10}\\
    \midrule
        H3DE \cite{huang2025H3DE} & 3.61$\pm$2.74 & 23.33 & 53.33 & 74.67 \\
        SR-UNet \cite{zhang2024SRLD} & 3.41$\pm$2.73 & 30.00 & 54.67 & 68.67 \\
        landmarker \cite{Jonkers2025landmarker} & 3.26$\pm$3.76 & 54.00 & 73.33 & 78.00 \\
        \midrule
        nnLandmark H3DE & 2.71$\pm$3.09 & 59.33 & 78.00 & 83.33 \\
        \midrule
        nnLandmark & 2.87$\pm$3.19 & 58.67 & 73.33 & 80.00 \\
        \textbf{nnLandmark ResEncM} & \textbf{4.03$\pm$10.14} & \textbf{56.67} & \textbf{72.00} & \textbf{79.33} \\
        nnLandmark ResEncL & 3.67$\pm$8.13 & 58.00 & 73.33 & 80.00 \\
    \bottomrule
    \multicolumn{5}{l}{\textbf{LFC} \#samples 60, \#landmarks 12}\\
    \midrule
         H3DE \cite{huang2025H3DE} & 4.22$\pm$4.21 & 34.03 & 55.42 & 67.78 \\
         SR-UNet \cite{zhang2024SRLD} & 3.92$\pm$3.62 & 35.42 & 55.97 & 69.72 \\
         landmarker \cite{Jonkers2025landmarker} & 4.02$\pm$5.13 & 50.56 & 68.19 & 76.67 \\
         \midrule
         nnLandmark H3DE & 3.72$\pm$4.47 & 51.94 & 67.91 & 77.22 \\
         \midrule
         nnLandmark & 3.72$\pm$4.47 & 51.94 & 67.91 & 77.22 \\
         \textbf{nnLandmark ResEncM} & \textbf{3.75$\pm$4.77} & \textbf{55.42} & \textbf{69.17} & \textbf{77.08} \\
         nnLandmark ResEncL & 3.75$\pm$4.75 & 54.44 & 68.61 & 76.53 \\
    \bottomrule
\end{tabular}
\end{table}

\section{Individual Landmark Class Errors}
\label{sec:app2_individual_mre}

\begin{table}[H]
\centering
\floatconts
    {tab:annex_res_afids}%
    {\caption{Landmark localization results of nnLandmark ResEncM for AFIDs dataset with \#samples 22, \#landmarks 32.}}%
    {\setlength{\tabcolsep}{3pt}}%
    \begin{tabular}{l c}
    \toprule
    \textbf{Landmark Class} & \textbf{MRE$\pm$Std} \\
    \midrule
        AC [midline] & 0.68$\pm$0.34\\
        PC [midline] & 0.90$\pm$0.39\\
        Infracollicular sulcus [midline] & 1.19$\pm$0.39\\
        Pontomesencephalic junction [midline] & 1.38$\pm$0.73\\
        Superior interpeduncular fossa [midline] & 0.87$\pm$0.35\\
        Right superior lateral mesencephalic sulcus & 1.19$\pm$0.50\\
        Left superior lateral mesencephalic sulcus & 1.06$\pm$0.54\\
        Right inferior lateral mesencephalic sulcus & 1.43$\pm$0.67\\
        Left inferior lateral mesencephalic sulcus & 1.35$\pm$0.68\\
        Culmen [midline] & 1.79$\pm$0.95\\
        Intermammillary sulcus [midline] & 1.10$\pm$0.46\\
        Right mammillary body & 0.95$\pm$0.42\\
        Left mamillary body & 1.12$\pm$0.50\\
        Pineal gland [midline] & 1.61$\pm$0.84\\
        Right lateral aspect of frontal horn at AC & 1.70$\pm$1.13\\
        Left lateral aspect of frontal horn at AC & 1.95$\pm$1.19\\
        Right lateral aspect of frontal horn at PC & 1.86$\pm$0.92\\
        Left lateral aspect of frontal horn at PC & 1.71$\pm$0.93\\
        Genu of corpus callosum [midline] & 1.10$\pm$0.39\\
        Splenium of the corpus callosum [midline] & 1.22$\pm$0.40\\
        Right anterolateral temporal horn & 1.16$\pm$0.70\\
        Left anterolateral temporal horn & 1.45$\pm$0.52\\
        Right superior AM temporal horn & 1.45$\pm$0.62\\
        Left superior AM temporal horn & 1.95$\pm$0.93\\
        Right inferior AM temporal horn & 2.14$\pm$0.91\\
        Left inferior AM temporal horn & 2.20$\pm$1.08\\
        Right indusium griseum origin & 1.42$\pm$0.78\\
        Left indusium griseum origin & 1.75$\pm$0.64\\
        Right ventral occipital horn & 1.80$\pm$1.60\\
        Left ventral occipital horn & 2.46$\pm$3.03\\
        Right olfactory sulcal fundus & 1.43$\pm$0.65\\
        Left olfactory sulcal fundus & 1.20$\pm$0.45\\
    \bottomrule
\end{tabular}
\end{table}

\begin{table}[H]
\centering
\floatconts
    {tab:annex_res_mml}%
    {\caption{Landmark localization results of nnLandmark ResEncM for MML dataset with \#samples 60, \#landmarks 14.}}%
    {\setlength{\tabcolsep}{3pt}}%
    \begin{tabular}{l c}
    \toprule
    \textbf{Landmark Class} & \textbf{MRE$\pm$Std} \\
    \midrule
        Left cuspid cusp & 1.29$\pm$1.06 \\
        Left 2nd molar crown & 0.96$\pm$0.56 \\
        Left 2nd molar mesial root & 1.26$\pm$0.65 \\
        Left 2nd molar distal root & 1.33$\pm$0.75 \\
        Left 3rd molar crown & 1.20$\pm$0.68 \\
        Left 3rd molar mesial root & 1.63$\pm$1.31 \\
        Left 3rd molar distal root & 1.90$\pm$0.89 \\
        Right cuspid cusp & 1.29$\pm$0.77 \\
        Right 2nd molar crown & 1.17$\pm$0.61 \\
        Right 2nd molar mesial root & 1.25$\pm$0.61 \\
        Right 2nd molar distal root & 1.41$\pm$0.79 \\
        Right 3rd molar crown & 1.04$\pm$0.46 \\
        Right 3rd molar mesial root & 1.50$\pm$0.81 \\
        Right 3rd molar distal root & 1.74$\pm$1.28 \\
    \bottomrule
\end{tabular}
\end{table}

\begin{table}[H]
\centering
\floatconts
    {tab:annex_res_pddca}%
    {\caption{Landmark localization results of nnLandmark ResEncM for PDDCA dataset with \#samples 7, \#landmarks 5.}}%
    {\setlength{\tabcolsep}{3pt}}%
    \begin{tabular}{l c}
    \toprule
    \textbf{Landmark Class} & \textbf{MRE$\pm$Std} \\
    \midrule
        Chin & 2.03$\pm$1.02 \\
        Left mandibular & 2.51$\pm$1.65 \\
        Right mandibular & 1.89$\pm$0.91 \\
        Occipital bone & 6.17$\pm$5.74 \\
        Odontoid process & 1.51$\pm$1.08 \\
    \bottomrule
\end{tabular}
\end{table}

\begin{table}[H]
\centering
\floatconts
    {tab:annex_res_fpose}%
    {\caption{Landmark localization results of nnLandmark ResEncM for fetal pose estimation dataset with \#samples 200, \#landmarks 22.}}%
    {\setlength{\tabcolsep}{3pt}}%
    \begin{tabular}{l c}
    \toprule
    \textbf{Landmark Class} & \textbf{MRE$\pm$Std} \\
    \midrule
        Cranial crest & 5.15$\pm$8.43\\
        Diencephalon & 2.56$\pm$4.74\\
        Thalamus & 1.84$\pm$1.32\\
        Nasal bone & 1.93$\pm$2.57\\
        Lower alveolar & 1.47$\pm$1.09\\
        Hind neck & 3.78$\pm$3.70\\
        Chest wall & 3.85$\pm$3.04\\
        Diaphragm lumbar & 3.71$\pm$4.10\\
        Buttocks & 3.98$\pm$5.91\\
        Umbilical & 4.09$\pm$4.41\\
        Left shoulder & 2.15$\pm$1.99\\
        Left elbow & 2.37$\pm$3.35\\
        Left wrist & 3.00$\pm$4.86\\
        Right shoulder & 2.99$\pm$5.39\\
        Right elbow & 2.54$\pm$4.00\\
        Right wrist & 3.13$\pm$4.95\\
        Left hip & 2.37$\pm$3.90\\
        Left knee & 2.60$\pm$4.10\\
        Left ankle & 3.96$\pm$5.24\\
        Right hip & 2.75$\pm$5.22\\
        Right knee & 2.55$\pm$2.99\\
        Right ankle & 4.51$\pm$5.26\\
    \bottomrule
\end{tabular}
\end{table}

\begin{table}[H]
\centering
\floatconts
    {tab:annex_res_feta}%
    {\caption{Landmark localization results of nnLandmark ResEncM for FeTA dataset with \#samples 15, \#landmarks 10.}}%
    {\setlength{\tabcolsep}{3pt}}%
    \begin{tabular}{l c}
    \toprule
    \textbf{Landmark Class} & \textbf{MRE$\pm$Std} \\
    \midrule
        Brain biparietal diameter 1 (bBIP1) & 0.82$\pm$0.56 \\
        Brain biparietal diameter 2 (bBIP2) & 1.41$\pm$0.53 \\
        Skull biparietal diameter 1 (sBIP1) & 4.67$\pm$5.78 \\
        Skull biparietal diameter 2 (sBIP2) & 4.46$\pm$5.35 \\
        Height of vermis 1 (HV1) & 3.50$\pm$2.90 \\
        Height of vermis 2 (HV2) & 10.04$\pm$20.14 \\
        Length of the corpus callosum 1 (LCC2) & 2.86$\pm$1.31 \\
        Length of the corpus callosum 2 (LCC2) & 9.47$\pm$21.23 \\
        Transverse cerebellar diameter 1 (TCD1) & 1.31$\pm$0.78 \\
        Transverse cerebellar diameter 2 (TCD2) & 1.75$\pm$1.28 \\
    \bottomrule
\end{tabular}
\end{table}

\begin{table}[H]
\centering
\floatconts
    {tab:annex_res_feta_biom}%
    {\caption{Biometry measurements results of nnLandmark ResEncM for FeTA with \#samples 15, \#measurements 10.}}%
    {\setlength{\tabcolsep}{3pt}}%
    \begin{tabular}{l c}
    \toprule
    \textbf{Landmark Class} & \textbf{MRE$\pm$Std} \\
    \midrule
        Brain biparietal diameter (bBIP) in the axial plane & 1.71$\pm$1.05 \\
        Skull biparietal diameter (sBIP) in the axial plane & 2.22$\pm$3.25 \\
        Height of the vermis (HV) in the sagittal plane & 2.47$\pm$2.67 \\
        Length of the corpus callosum (LCC) in the sagittal plane & 1.66$\pm$1.94 \\
        Maximum transverse cerebellar diameter (TCD) in the coronal plane & 2.06$\pm$3.37 \\
    \bottomrule
\end{tabular}
\end{table}

\begin{table}[H]
\centering
\floatconts
    {tab:annex_res_lfc}%
    {\caption{Landmark localization results of nnLandmark ResEncM for LFC dataset with \#samples 60, \#landmarks 12.}}%
    {\setlength{\tabcolsep}{3pt}}%
    \begin{tabular}{l c}
    \toprule
    \textbf{Landmark Class} & \textbf{MRE$\pm$Std} \\
    \midrule
        Brain biparietal diameter 1 (bBIP1) & 6.09$\pm$5.76 \\
        Brain biparietal diameter 2 (bBIP2) & 6.15$\pm$5.74 \\
        Skull biparietal diameter 1 (sBIP1) & 6.11$\pm$4.77 \\
        Skull biparietal diameter 2 (sBIP2) & 5.99$\pm$4.97 \\
        Transverse cerebellar diameter 1 (TCD1) & 1.30$\pm$0.64 \\
        Transverse cerebellar diameter 2 (TCD2) & 1.21$\pm$0.69 \\
        Occipitofrontal diameter 1 (OFD1) & 6.52$\pm$7.32 \\
        Occipitofrontal diameter 2 (OFD2) & 6.09$\pm$5.83 \\
        Height of vermis 1 (HDV1) & 1.41$\pm$0.66 \\
        Height of vermis 2 (HDV2) & 1.45$\pm$0.58 \\
        Anteroposterior diameter of vermis 1 (ADV1) & 1.57$\pm$0.68 \\
        Anteroposterior diameter of vermis 2 (ADV2) & 1.14$\pm$0.55 \\
    \bottomrule
\end{tabular}
\end{table}

\begin{table}[H]
\centering
\floatconts
    {tab:annex_res_lfc_biom}%
    {\caption{Biometry measurements results of nnLandmark ResEncM for LFC with \#samples 60, \#measurements 6.}}%
    {\setlength{\tabcolsep}{3pt}}%
    \begin{tabular}{l c}
    \toprule
    \textbf{Landmark Class} & \textbf{MRE$\pm$Std} \\
    \midrule
        Brain biparietal diameter (bBIP) & 1.49$\pm$1.23 \\
        Skull biparietal diameter (sBIP) & 1.24$\pm$0.94 \\
        Transverse cerebellar diameter (TCD) & 1.07$\pm$0.76 \\
        Occipitofrontal diameter (OFD) & 1.25$\pm$0.94 \\
        Height of vermis (HDV) & 1.30$\pm$0.84 \\
        Anteroposterior diameter of vermis (ADV) & 0.85$\pm$0.65 \\
    \bottomrule
\end{tabular}
\end{table}

\clearpage
\section{Reproducibility Results}
\label{sec:app3_reproducibility}

\begin{table}[H]
\centering
\floatconts
    {tab:annex_reproducibility}%
    {\caption{Reproducibility results on the complete subset of the Mandibular Molar Landmark (MML) dataset.}}%
    {\setlength{\tabcolsep}{3pt}}%
    \begin{tabular}{l c}
    \toprule
    \textbf{Method} & \textbf{MRE$\pm$Std [mm]} \\
    \midrule
        H3DE reported \cite{huang2025H3DE} & 1.68$\pm$0.45 \\
        H3DE reproduced \cite{huang2025H3DE} & 1.81$\pm$1.15 \\
        SR-UNet reported \cite{zhang2024SRLD} & 2.01$\pm$4.33 \\
        SR-UNet reproduced \cite{zhang2024SRLD} & 10.01$\pm$10.37 \\
        landmarker reported (different split) \cite{Jonkers2025landmarker} & 1.39 \\
        landmarker reproduced \cite{Jonkers2025landmarker} & 10.58$\pm$13.92 \\
    \bottomrule
\end{tabular}
\end{table}

\clearpage
\section{Qualitative Examples per Dataset}
\label{sec:app4_qualitative}

\begin{figure}[H]
    \centering
    \includegraphics[width=430pt]{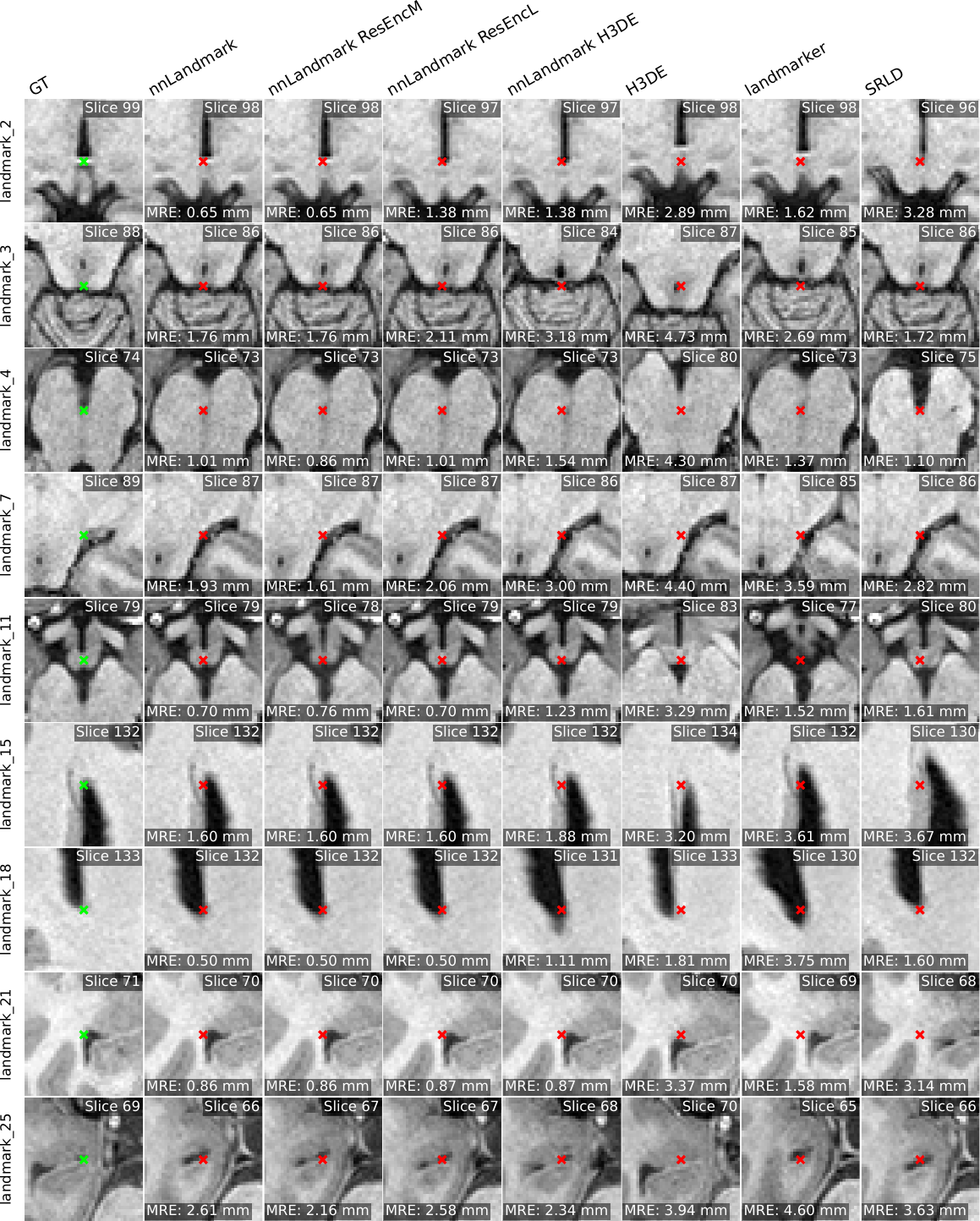}
    \caption{Qualitative examples of selected landmarks on Afids dataset.} 
    \label{fig3}
\end{figure}

\begin{figure}[t]
    \centering
    \includegraphics[width=430pt]{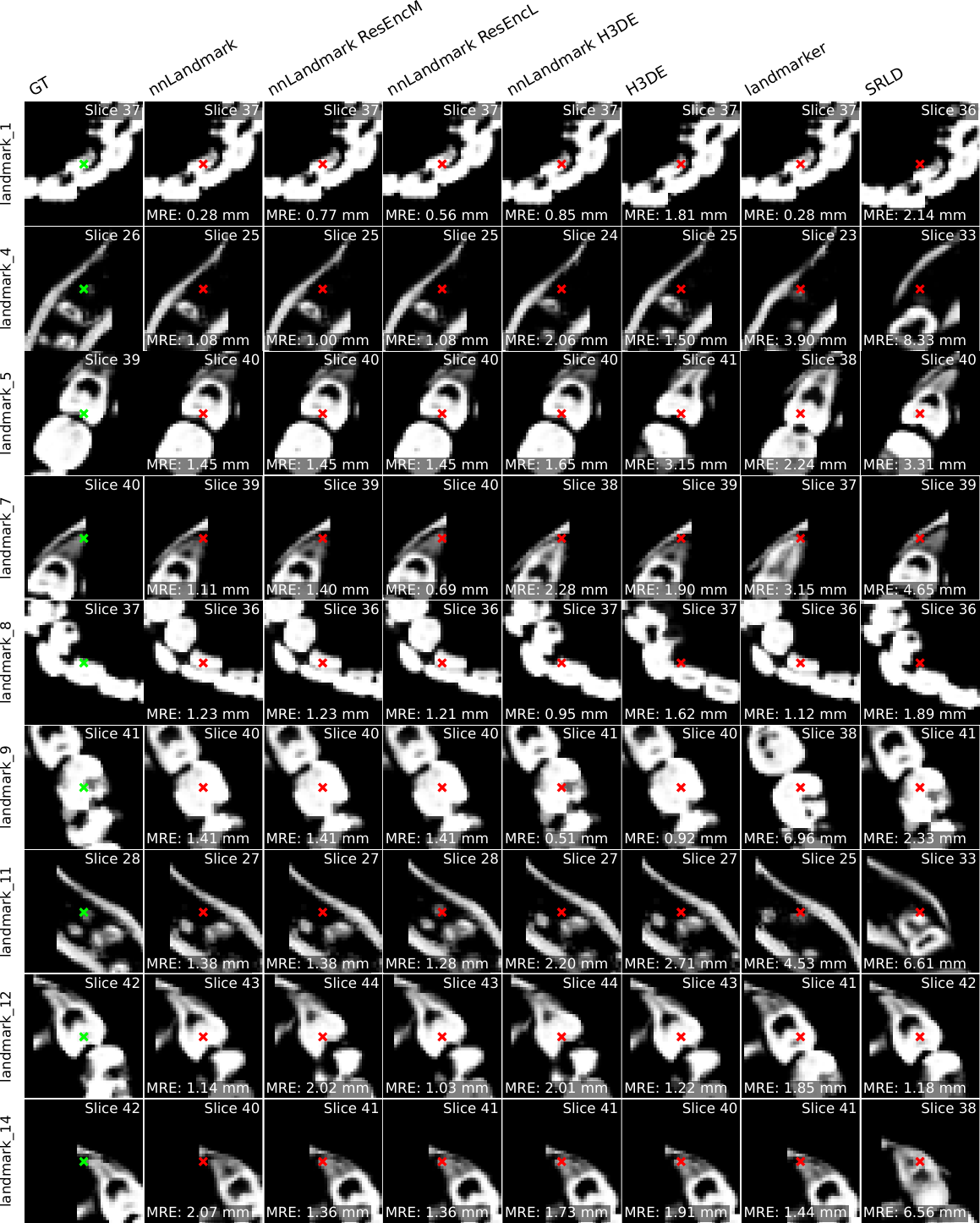}
    \caption{Qualitative examples of selected landmarks on MML dataset.} 
    \label{fig3}
\end{figure}

\begin{figure}[t]
    \centering
    \includegraphics[width=430pt]{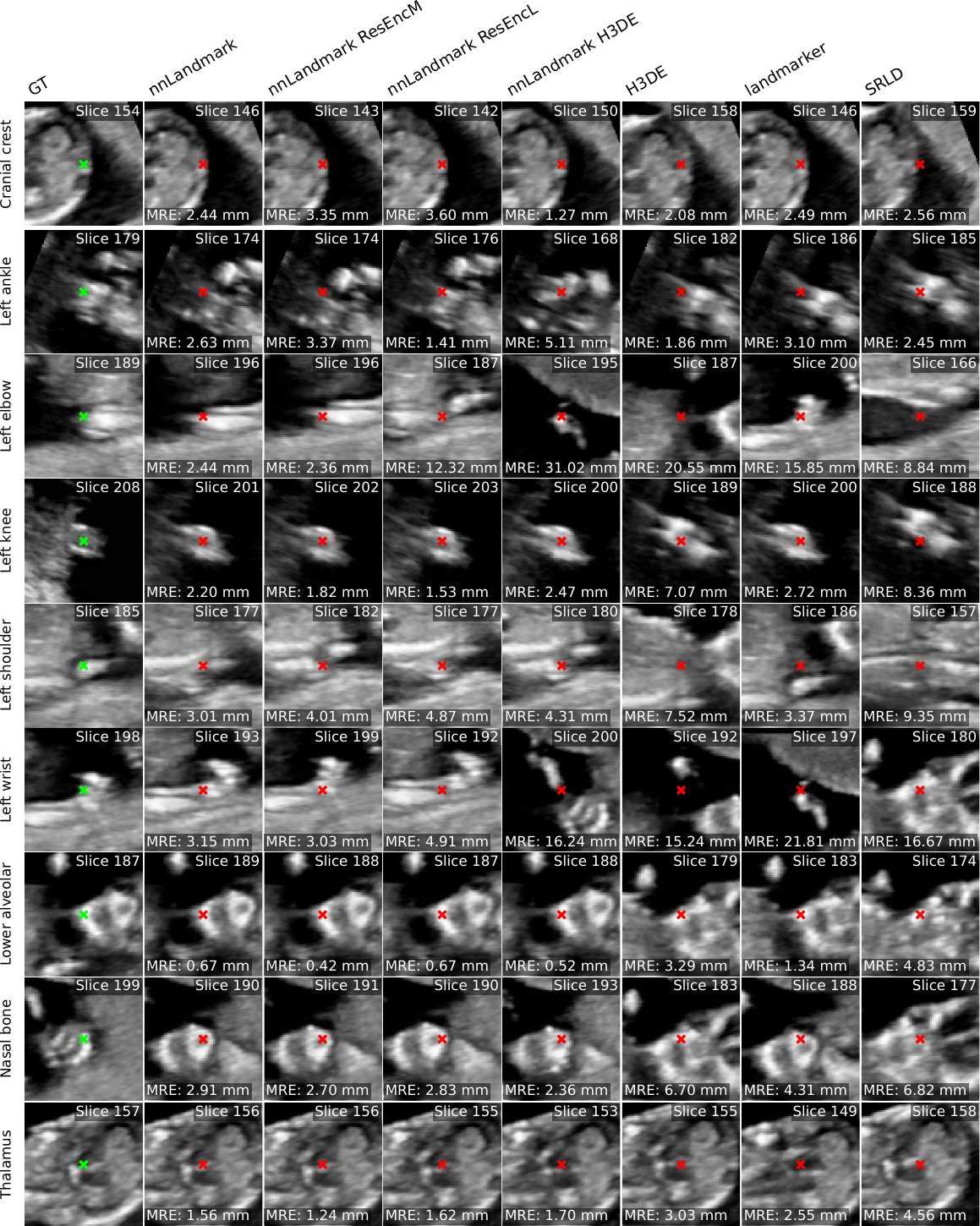}
    \caption{Qualitative examples of selected landmarks on Fetal pose dataset.} 
    \label{fig3}
\end{figure}

\begin{figure}[t]
    \centering
    \includegraphics[width=430pt]{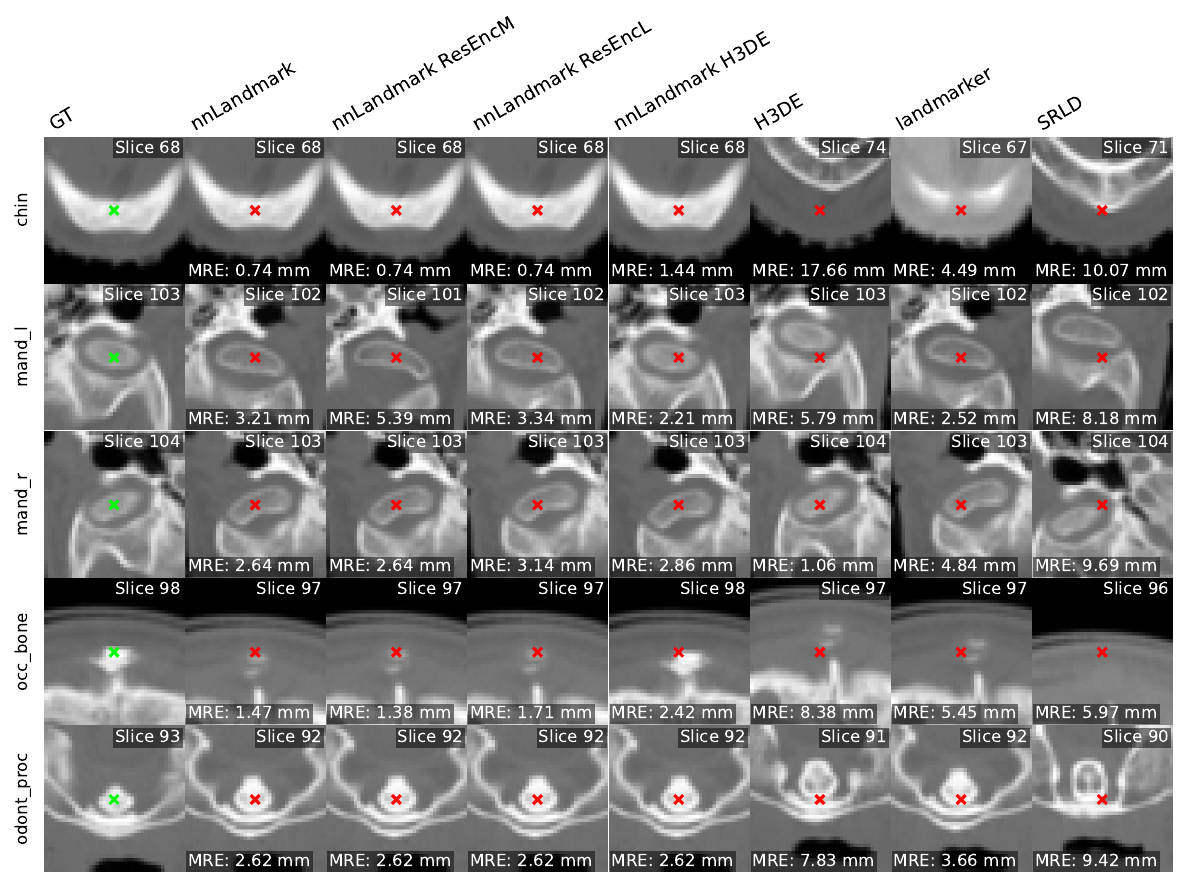}
    \caption{Qualitative examples of selected landmarks on PDDCA dataset.} 
    \label{fig3}
\end{figure}

\begin{figure}[t]
    \centering
    \includegraphics[width=430pt]{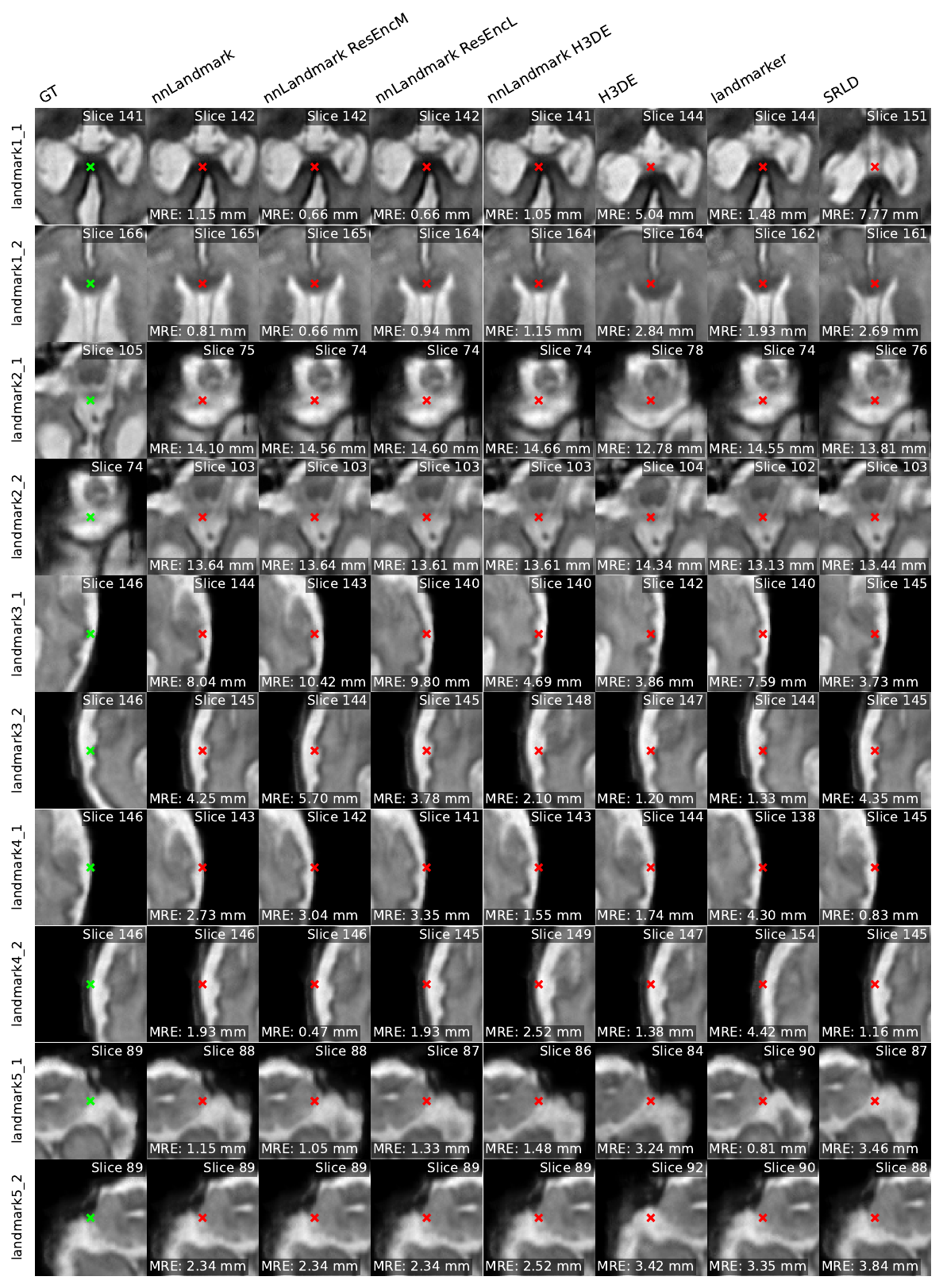}
    \caption{Qualitative examples of selected landmarks on FeTA dataset.} 
    \label{fig3}
\end{figure}

\begin{figure}[t]
    \centering
    \includegraphics[width=430pt]{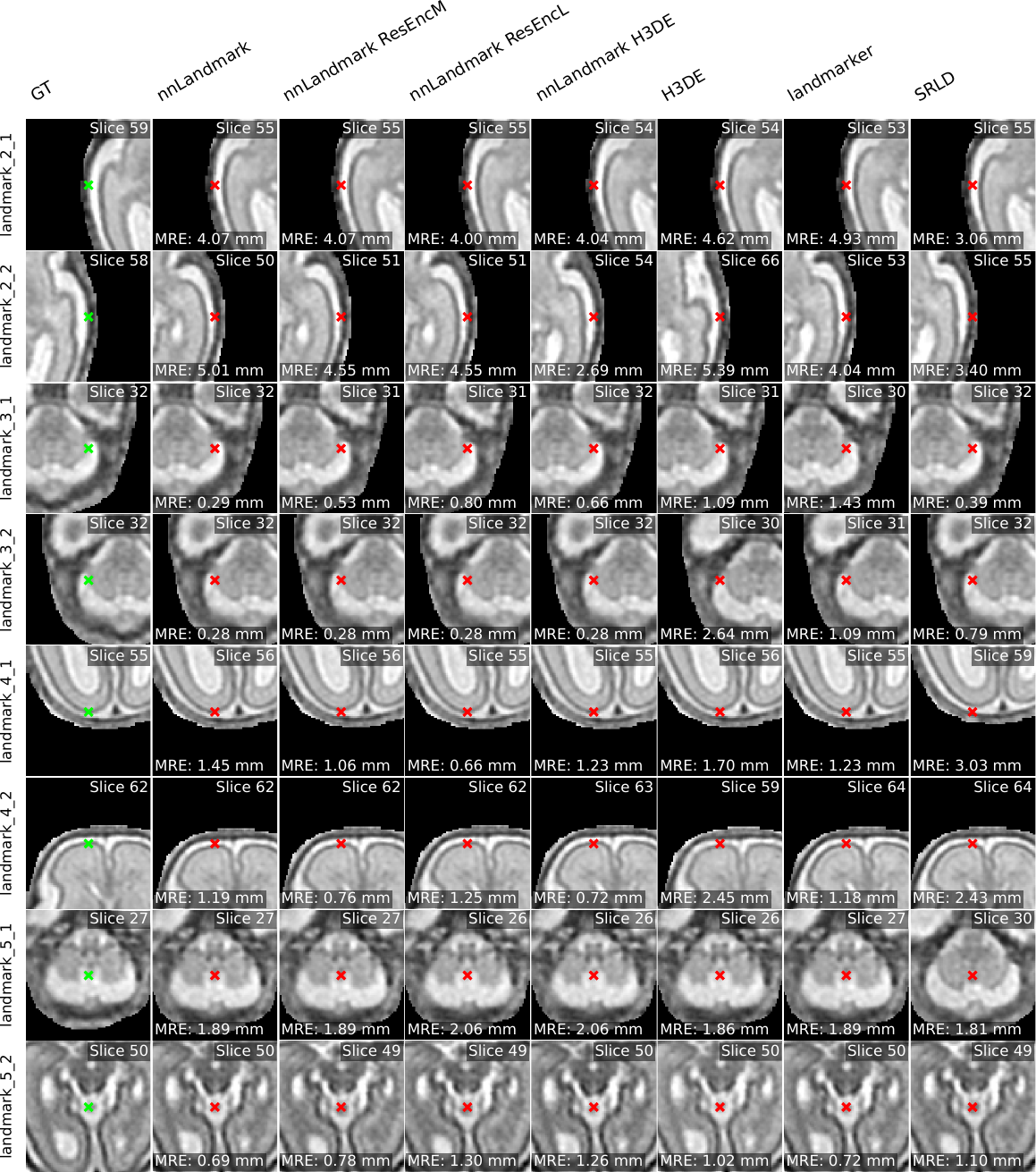}
    \caption{Qualitative examples of selected landmarks on LFC dataset.} 
    \label{fig3}
\end{figure}

\end{document}